\documentclass[10pt,twocolumn,letterpaper]{article}

\usepackage[]{cvpr} 

\usepackage{subcaption}
\usepackage{caption}
\usepackage{adjustbox}
\usepackage{amssymb}
\usepackage{multirow}
\usepackage{booktabs}
\usepackage{xcolor}
\usepackage[normalem]{ulem}
\usepackage{colortbl}
\newcommand{\gc}[1]{\textcolor{green!60!black}{#1}} 
\newcommand{\rc}[1]{\textcolor{red!70!black}{#1}}   
\newcommand{\bc}[1]{\textcolor{cyan!60!black}{#1}}  
\usepackage{booktabs, tabularx, array}
\usepackage{makecell}
\usepackage{cuted}
\usepackage{stfloats}

\usepackage{array}
\newcolumntype{R}[1]{>{\raggedleft\arraybackslash}m{#1}} 
\newcolumntype{C}[1]{>{\centering\arraybackslash}m{#1}} 
\usepackage{algorithm}
\usepackage{algorithmic}
\newcolumntype{Y}{>{\centering\arraybackslash}X}

\definecolor{cvprblue}{rgb}{0.21,0.49,0.74}
\newcommand{\cvprref}[1]{{\color{cvprblue}#1}}
\usepackage[pagebackref,breaklinks,colorlinks,allcolors=cvprblue]{hyperref}
\usepackage{amsmath}
\def\confName{CVPR}
\def\confYear{2026}

\usepackage[
   enable,
   labname={CMLab},
   logo={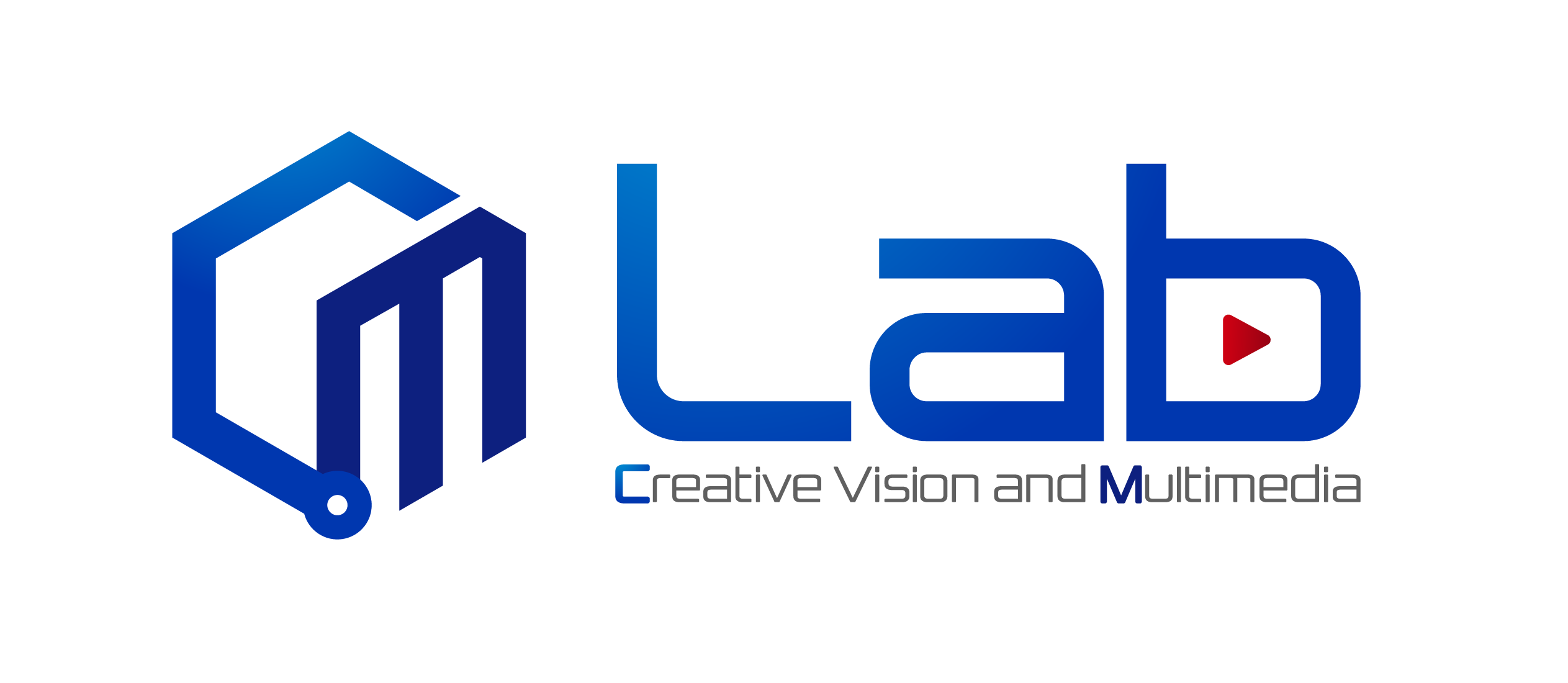},
   logowidth={26mm},
   projectpage={https://cmlab-korea.github.io/FRAMER/}
]{cmlab}

\title{\includegraphics[scale=0.5, keepaspectratio]{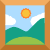} FRAMER: Frequency-Aligned Self-Distillation with Adaptive Modulation Leveraging Diffusion Priors for Real-World Image Super-Resolution}

\author{
Seungho Choi\hspace{2em} Jeahun Sung\hspace{2em} Jihyong Oh$^{\dagger}$\\[0.5em]
Creative Vision and Multimedia Lab (CMLab)\\
Chung-Ang University, Seoul, South Korea\\[0.2em]
{\tt\small \{choiseungho1019, jhseong, jihyongoh\}@cau.ac.kr}\\
{\small{\url{https://cmlab-korea.github.io/FRAMER/}}}
}

\cmlabAuthors{Seungho Choi \qquad Jeahun Sung \qquad Jihyong Oh$^{\dagger}$}
\cmlabAuthorsFootnotes{$^{\dagger}$ Corresponding author}
\cmlabAffiliations{CMLab, Chung-Ang University}
\cmlabAuthorEmail{\small \{choiseungho1019, jhseong, jihyongoh\}@cau.ac.kr}
\cmlabProjectPage{https://cmlab-korea.github.io/FRAMER/}

\begin{document}

\maketitle
\footnotetext{$^{\dagger}$Corresponding author}
\begin{strip}
\begin{minipage}{\textwidth}
\vspace{-12mm}
\centering
\includegraphics[width=\linewidth]{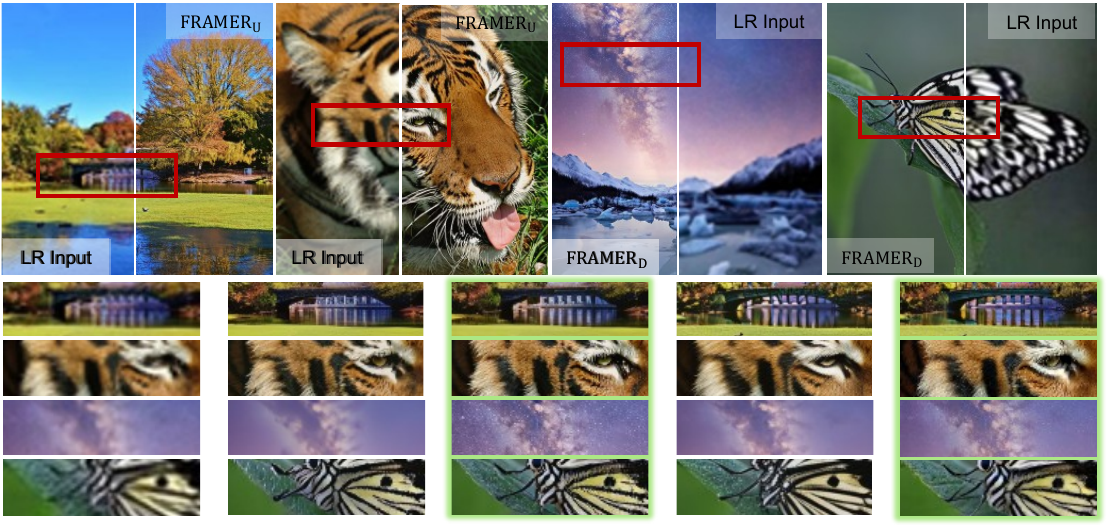}
\put(-471.0, -11.0){LR Input}
\put(-376.0, -11.0){DiT4SR~\cite{duan2025dit4sr}}
\put(-275.0, -11.0){FRAMER$_D$}
\put(-175.0, -11.0){PiSA-SR~\cite{sun2025pixel}}
\put(-70.0, -11.0){FRAMER$_U$}
\vspace{-2mm}
\captionof{figure}{\textbf{Qualitative comparison with recent Real-ISR methods on real-world images.}
Our FRAMER models produce sharper edges and richer details, leading to more visually natural and faithful restoration results.
More qualitative results are provided in Supplementary.
}
\label{fig:qualitative_comparisons}
\vspace{-2mm}
\end{minipage}
\end{strip}

\begin{abstract}
Real-image super-resolution (Real-ISR) seeks to recover HR images from LR inputs with mixed, unknown degradations. While diffusion models surpass GANs in perceptual quality, they under-reconstruct high-frequency (HF) details due to a low-frequency (LF) bias and a depth-wise “low-first, high-later” hierarchy. We introduce \textbf{FRAMER}, a plug-and-play training scheme that exploits diffusion priors without changing the backbone or inference. At each denoising step, the final-layer feature map teaches all intermediate layers. Teacher and student feature maps are decomposed into LF/HF bands via FFT masks to align supervision with the model’s internal frequency hierarchy. For LF, an Intra Contrastive Loss (IntraCL) stabilizes globally shared structure. For HF, an Inter Contrastive Loss (InterCL) sharpens instance-specific details using random-layer and in-batch negatives. Two adaptive modulators, Frequency-based Adaptive Weight (FAW) and Frequency-based Alignment Modulation (FAM), reweight per-layer LF/HF signals and gate distillation by current similarity. Across U-Net and DiT backbones (e.g., Stable Diffusion 2, 3), FRAMER consistently improves PSNR/SSIM and perceptual metrics (LPIPS, NIQE, MANIQA, MUSIQ). Ablations validate the final-layer teacher and random-layer negatives. 
\end{abstract}    
\section{Introduction}
\label{sec:introduction}

Real Image Super-Resolution (Real-ISR)~\cite{wang2021real} aims to recover a high-resolution (HR) image from a low-resolution (LR) counterpart suffering from mixed, unknown degradations (e.g., blur, noise)~\cite{wang2021real}. Unlike traditional SR~\cite{dong2014learning}, Real-ISR must simultaneously remove complex degradations and generate realistic details, demanding stronger model capacity and robustness~\cite{wu2024seesr,10.1007/978-3-031-73202-7_25}.

While SR methods evolved from early CNNs to GANs~\cite{10.1016/j.neucom.2024.128911}, GANs often suffer from training instability under the complex degradations of Real-ISR~\cite{kuznedelev2024does,10.1145/3696271.3696307}. Diffusion models have surpassed them by offering stable training and superior perceptual quality~\cite{dhariwal2021diffusion,saharia2022image}. Leveraging large-scale pre-trained text-to-image (T2I) models (e.g., Stable Diffusion with U-Net~\cite{ronneberger2015u} or DiT~\cite{peebles2023scalable} backbones) is a promising direction due to their rich, real-world priors~\cite{SD2,peebles2023scalable,zhang2018unreasonable}. However, these priors are optimized for creative generation, not the high-fidelity, task-constrained demands of Real-ISR~\cite{wu2024seesr,wan2024controlsr}. The key challenge is adapting these priors for our task.

\begin{figure}
    \centering
    \includegraphics[width=1.0\linewidth]{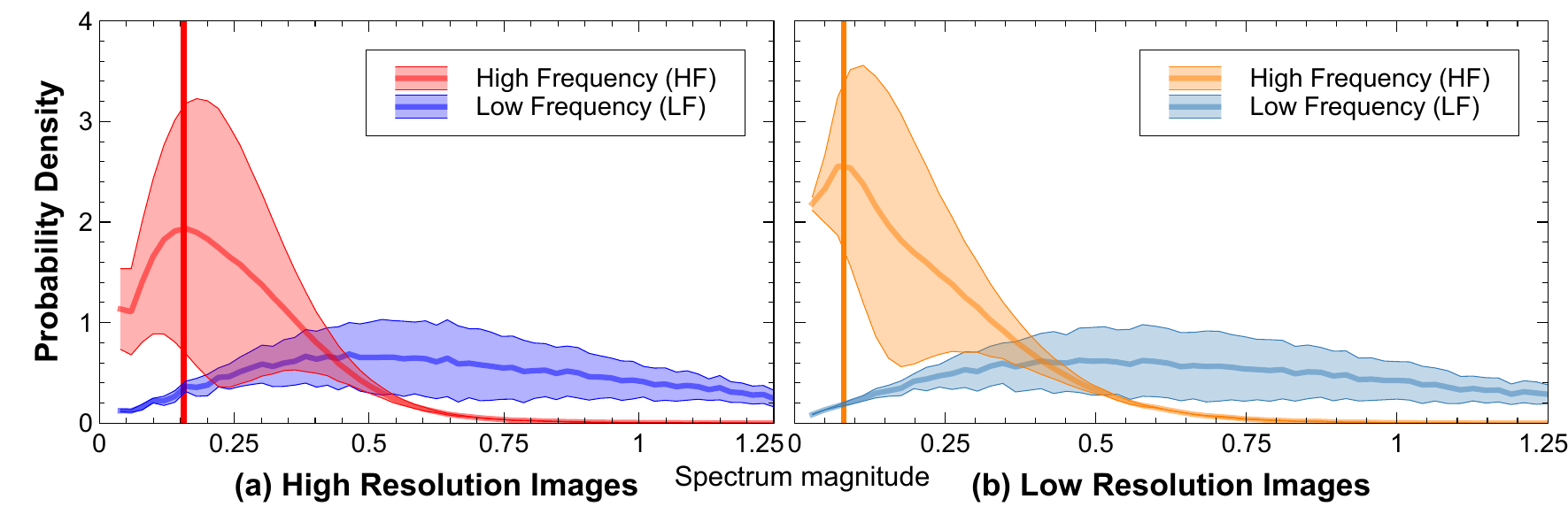}
    \caption{\textbf{Band-wise magnitude densities with shared bins.} For each feature map, we compute the 2D FFT and collect magnitudes $\lvert F\rvert$ within LF and HF rings. We plot mean $\pm$ $\sigma$ densities over samples for $\log(1+\lvert F\rvert)$ using common bin edges (HF: red or yellow, LF: blues). LF magnitudes span a broader and heavier range, whereas HF magnitudes concentrate narrowly near small values, indicating LF dominance that biases unified training toward LF and undertrains HF details. All statistics are computed on the 100-image DIV2K~\cite{agustsson2017ntire} test set. Densities integrate to 1; any right-edge spike is due to percentile clipping used only for visualization.}
    \label{fig:band_hist}
\end{figure}

\begin{figure}
    \centering
    \includegraphics[width=\linewidth]{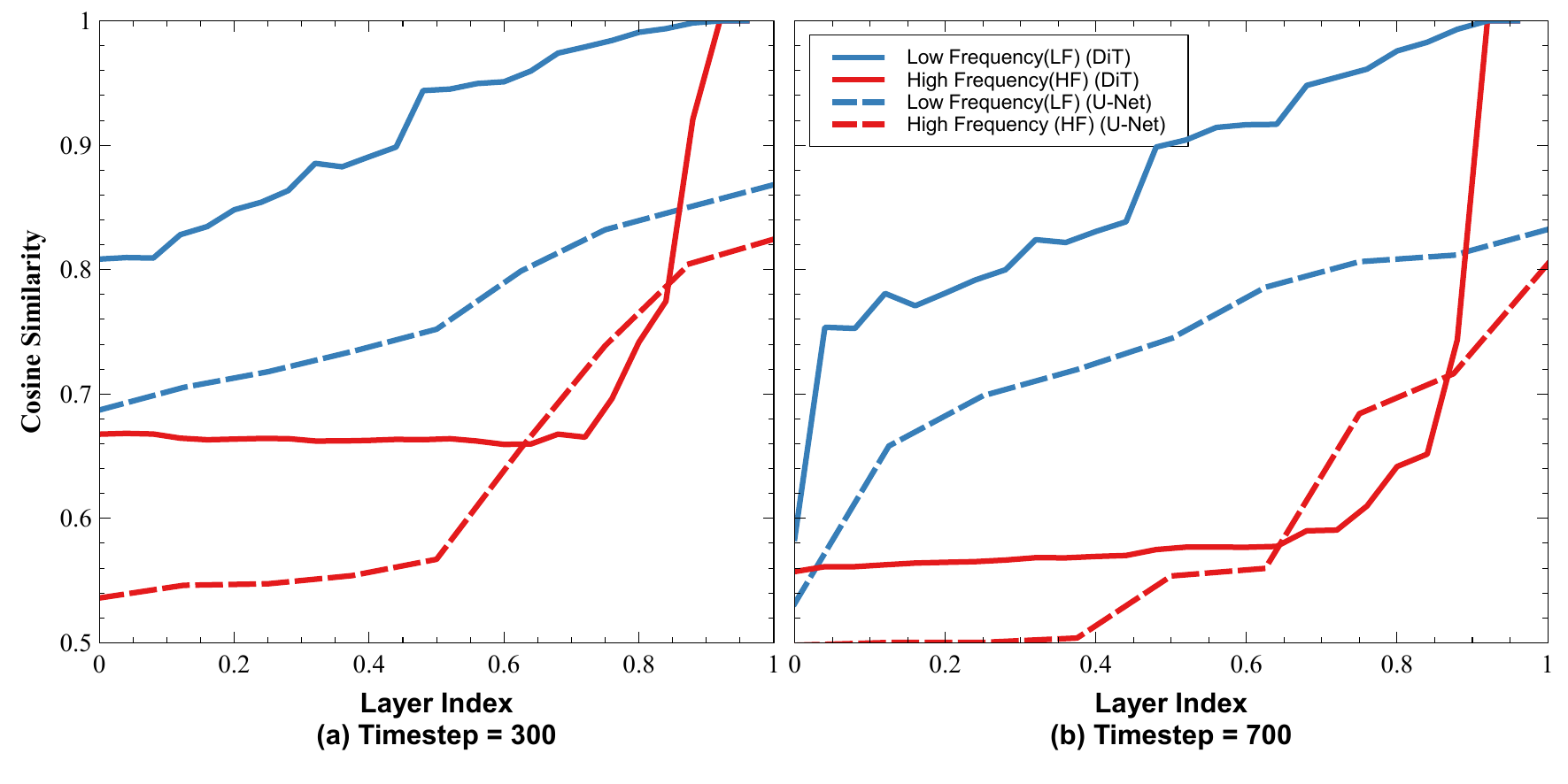}
    \caption{
    \textbf{Layer-wise cosine similarity of LF and HF feature maps in U-Net~\cite{ronneberger2015u} (dotted line) and DiT~\cite{peebles2023scalable} (solid line).} (a) low-noise timestep ($t{=}300$), (b) high-noise timestep ($t{=}700$).Using the final-layer feature map as reference, LF similarity converges faster in earlier layers, whereas HF similarity rises abruptly in later layers. This reveals a “low-first, high-later” depth-wise hierarchy (i.e., an LF bias), motivating our layer-adaptive, frequency-aware training strategy. (For comparability, layer depth is normalized to [0, 1].)}
    
    \label{fig:layer-wise-cos}
\end{figure}

Despite their success, diffusion models still struggle to reconstruct fine high-frequency (HF) details, often yielding over-smoothed results~\cite{medi2025missing}. We trace this to a fundamental low-frequency (LF) bias stemming from two properties. First, natural image frequency distributions are inherently LF-dominant, an imbalance that is severely exacerbated in the LR inputs (\refig{fig:band_hist}). The standard noise-prediction loss~\cite{ho2020denoising} thus favors these dominant LF components to reduce overall loss, inevitably undertraining HF signals~\cite{choi2022perception,falck2025fourier}. Second, we identify a depth-wise ``low-first, high-later'' hierarchy. While not a strict universal rule, this is a broadly observed empirical trend in diffusion-based restoration models: an analysis of layer-wise feature maps (\refig{fig:layer-wise-cos}) shows that LF features stabilize early in the network, while HF features converge only near the final layers. A conventional, frequency-agnostic loss is fundamentally misaligned with this hierarchy. It supplies redundant LF-biased gradients to early layers (where LF has already stabilized) while starving the later, HF-refining layers of the necessary optimization signals.

A simple approach, adding explicit frequency-domain losses, fails due to a domain mismatch between high-dimensional features and pixel-space targets~\cite{tian2020contrastive,zhao2022decoupled}. Self-distillation (SD)~\cite{hinton2015distilling}, however, offers a well-aligned alternative~\cite{SDSSL,SSSD}. We treat the final-layer feature map as the \textit{teacher} and intermediate layers as \textit{students}. This avoids domain mismatch, as both operate within the same feature space, providing a consistent and domain-compatible distillation signal~\cite{zhang2019your}.

However, standard SD is frequency-agnostic. To correct the identified spectral bias and layer-wise misalignment, we propose \textbf{Fr}equency-\textbf{A}ligned Self-Distillation with Adaptive \textbf{M}odulation Lev\textbf{er}aging Diffusion Priors, namely \textbf{FRAMER}, a layer-adaptive SD framework with four frequency-aware components. First, \textbf{Intra Contrastive Loss (IntraCL)} (\refsec{sec:intracl}) focuses on LF stability. Second, \textbf{Inter Contrastive Loss (InterCL)} (\refsec{sec:intercl}) targets HF detail sharpness. Third, \textbf{Frequency-based Adaptive Weight (FAW)} (\refsec{sec:faw}) adaptively modulates distillation strength across layers and frequencies. Finally, \textbf{Frequency-based Alignment Modulation (FAM)} (\refsec{sec:fam}) controls strength based on student-teacher similarity to prevent collapse and stabilize training.

In summary, our main contributions are:

\begin{itemize}
    \item We propose \textbf{FRAMER}, a self-distillation framework that introduces frequency-aligned contrastive losses, IntraCL for LF stability and InterCL for HF details, to counteract the LF bias in SR training tendency of the standard noise-prediction loss~\cite{ho2020denoising}.
    \item We further introduce adaptive distillation mechanisms, FAW and FAM, based on the internal frequency hierarchy. Together they suppress unstable gradients, prevent early-layer collapse and overfitting, accelerate convergence, resulting in improving SR quality in terms of diverse metrics with better training stability.
    \item FRAMER is a plug-and-play training technique, effectively leveraging the rich priors of pre-trained text-to-image models, such as Stable Diffusion 2~\cite{SD2} (U-Net) and Stable Diffusion 3~\cite{SD3} (DiT), without altering their architecture or inference.
\end{itemize}
\section{Related Work}

\noindent\textbf{Image Super-Resolution.} Early SR followed supervised regression under synthetic bicubic downsampling, then moved to blind SR for unknown degradations, and progressed to Real-ISR emphasizing realistic evaluation~\cite{wang2021real}. CNN-based methods improved distortion but were often oversmoothed; GANs increased realism but suffered from stability issues. Diffusion models advanced SR with stronger perceptual quality~\cite{dhariwal2021diffusion,saharia2022image}. Research then adopted pre-trained models like Stable Diffusion to leverage real-world priors~\cite{duan2025dit4sr,sun2025pixel,cheng2025effective}. However, these pipelines supervise all layers and frequencies with a single, standard noise-prediction loss, and thus do not exploit the early LF and late HF behavior along timesteps and depth. FRAMER resolves this gap with a novel self-distillation framework that is both frequency-aligned and layer-adaptive.

\noindent\textbf{Frequency-Aware Learning in Diffusion Models.} Several works shape the diffusion process to be frequency-aware. Approaches include manipulating forward noising in the frequency domain~\cite{jiralerspong2025shaping}, designing frequency-guided multi-scale diffusion with wavelets~\cite{wang2024frequency}, introducing HF-preserving objectives to align features~\cite{sami2024hf}, or using training-free inference-time modulation~\cite{li2025fedsr}. These methods increase frequency awareness but generally rely on fixed schedules or auxiliary modules. Critically, they do not adapt supervision to what each layer actually changes at a given timestep. FRAMER addresses this specific gap with its FAW, which modulates supervision based on the actual layer-wise change rate, and FAM, which gates distillation based on student-teacher alignment, rather than relying on a pre-defined, fixed schedule.

\begin{figure*}
    \centering
    \includegraphics[width=0.8\linewidth]{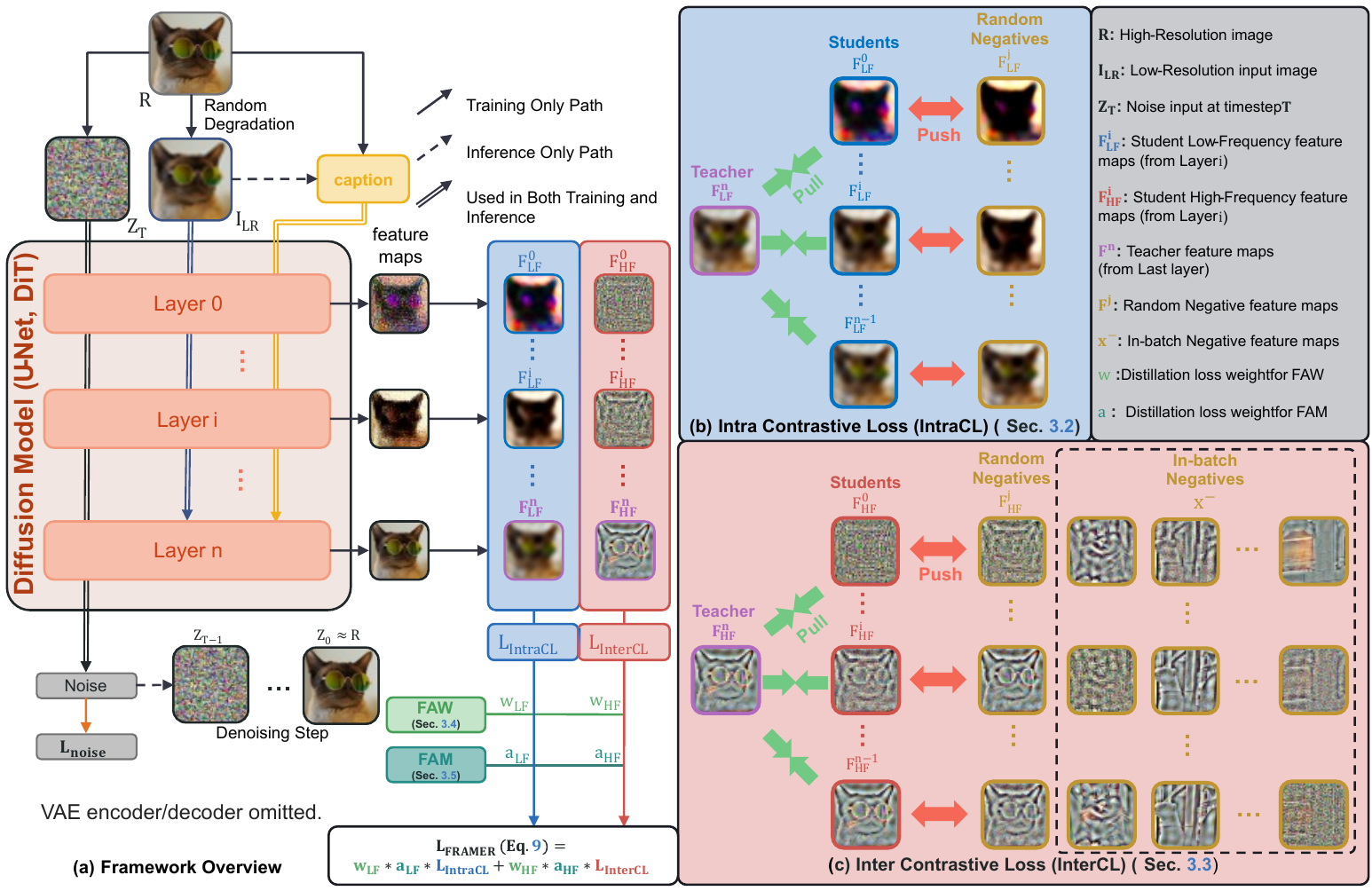}
    \caption{\textbf{FRAMER: \underline{Fr}equency-\underline{A}ligned Self-Distillation with Adaptive \underline{M}odulation Lev\underline{er}aging Diffusion Priors} (inspired by \refsec{sec:observations}).
\textbf{(a) Framework Overview.} During training, from an High-Resolution image \(R\), we create \(I_{LR}\) by random degradation~\cite{wang2021real}, downsampling, and resizing back to the size of \(R\). We use LLaVA~\cite{liu2023llava} to generate a caption. The diffusion backbone (U-Net~\cite{ronneberger2015u}/DiT~\cite{peebles2023scalable}) takes \(I_{LR}\), noise \(Z_T\), and the caption as inputs; FRAMER is applied only during training and jointly uses IntraCL (\refsec{sec:intracl}) and InterCL (\refsec{sec:intercl}), modulated by FAW (\refsec{sec:faw}) and FAM (\refsec{sec:fam}). FRAMER is a training framework that adds auxiliary loss components only during training. At inference, it uses the original backbone without any modification, making it fully plug-and-play.
\textbf{(b) Intra Contrastive Loss.} Within a single image, pull \(F^{(i)}_{\mathrm{LF}}\) toward \(F^{(n)}_{\mathrm{LF}}\) and push it away from a randomly sampled same-image layer \(F^{(j)}_{\mathrm{LF}}\) (no in-batch negatives).
\textbf{(c) Inter Contrastive Loss.} Attract \(F^{(i)}_{\mathrm{HF}}\) to \(F^{(n)}_{\mathrm{HF}}\) and repel a random-layer negative \(F^{(j)}_{\mathrm{HF}}\) (same image) and in-batch negatives \(x^{-}\) (other images).}
    \label{fig:main}
\end{figure*}

\noindent\textbf{Self-Distillation in Diffusion Models.} Self-distillation (SD) has been applied to improve generative fidelity beyond sampling acceleration. Methods include unified SD losses~\cite{woods2024selfdistillation}, step-level fusion aggregating predictions across timesteps~\cite{Ma_2025_CVPR}, and representation alignment enforcing feature consistency~\cite{jiang2025no,berrada2025boosting}. These techniques are powerful for aligning representations but remain fundamentally frequency-agnostic; they align the entire feature map, thus implicitly inheriting the LF-bias of standard losses. FRAMER complements this line by explicitly introducing frequency-awareness to the SD framework. By using a dedicated IntraCL for LF stability and an InterCL for HF detail, it separates the distillation objectives based on spectral roles, directly counteracting the LF-bias instead of inheriting it.

\section{Proposed Method}

Training diffusion models for Real-ISR faces two key challenges~\cite{choi2022perception,falck2025fourier}: (1) the noise-prediction loss is biased toward LF components, and (2) this loss is misaligned with the network's ``low-first, high-later'' processing, leaving HF layers under-optimized (\refig{fig:layer-wise-cos}).

Simple auxiliary frequency-based losses for intermediate layers fail, introducing a cross-domain objective mismatch~\cite{tian2020contrastive,zhao2022decoupled} that forces high-level features to regress to different-domain targets, destabilizing training~\cite{romero2014fitnets}.

To avoid this mismatch, we leverage priors from pre-trained diffusion backbones (e.g., SD2, SD3)~\cite{SD2,SD3}, which encode rich natural-image statistics. We realign these priors for Real-ISR via frequency-aware, layer-adaptive supervision without changing inference. We use self-distillation (SD)~\cite{hinton2015distilling,SDSSL,SSSD}: intermediate layers (\textit{students}) are supervised by the final-layer feature map (\textit{teacher})~\cite{ho2020denoising,SD2,peebles2023scalable}. This provides a domain-compatible signal, avoiding the cross-domain mismatch.

\subsection{Observations}
\label{sec:observations}

\textbf{(1) LF-dominant training statistics in Real-ISR.} In Real-ISR, HR/LR images have unbalanced band-wise magnitudes: LF is broad and heavy, while HF is narrow and small. Thus, the noise-prediction objective is LF-biased and undertrains HF~\cite{choi2022perception,falck2025fourier} (cf. \refig{fig:band_hist}).

\noindent\textbf{(2) Shared LF vs. instance-specific HF.} Per \refig{fig:in_batch_sims}, LF features show high cross-sample similarity (shared structures), while HF features show low similarity (instance-specific details). This implies in-batch negatives are false negatives for LF but informative for HF.

\noindent\textbf{(3) Depth-wise ``low-first, high-later'' hierarchy.} Consistent with \refig{fig:layer-wise-cos}, diffusion backbones show a ``low-first, high-later'' frequency-progressive hierarchy: LF stabilizes in early layers, while HF similarity rises in later layers.

\noindent\textbf{(4) Degrees of alignment increase along layer depth.} Per \refig{fig:layer-wise-cos}, student–teacher feature alignment is low in early layers and increases with depth.



\noindent\textbf{Overview.} Building on \refsec{sec:observations}, we propose \textbf{FRAMER} (\textbf{Fr}equency-\textbf{A}ligned Self-Distillation with Adaptive \textbf{M}odulation Lev\textbf{er}aging Diffusion Priors) (\refig{fig:main}). During training, we create LR inputs $I_{\mathrm{LR}}$ from HR images $R$ via random degradation~\cite{wang2021real} and get captions via LLaVA~\cite{liu2023llava}. The model is conditioned on $(I_{\mathrm{LR}}, Z_T, \text{caption})$. 

During training, the final-layer feature map (\textit{teacher}) supervises intermediate layers (\textit{students}). We decompose teacher/student features into LF/HF bands (see \refsec{sec:a1}, \refsec{sec:a2} for justification). We then: (i) apply IntraCL (\refsec{sec:intracl}) to the LF band to stabilize shared structure; (ii) apply InterCL (\refsec{sec:intercl}) to the HF band to sharpen instance-details; (iii) compute FAW (\refsec{sec:faw}) to adaptively weight distillation based on student–teacher discrepancy; and (iv) apply FAM (\refsec{sec:fam}) to gate the distillation by student–teacher alignment. These terms are added to the noise-prediction loss. The full algorithm is provided in Supplementary \refsec{sec:training_algo}.

In essence, FRAMER can be viewed as a synergistic combination of (i) frequency-targeted contrastive distillation (via IntraCL and InterCL) for balanced spectral optimization, and (ii) a unified adaptive stabilization mechanism (via FAW and FAM) that modulates the supervision strength according to the model’s internal hierarchical progression.
As a \textbf{strictly training-only} strategy, FRAMER introduces \textbf{no inference overhead} and incurs minimal training cost (\textit{Suppl.}, \refsec{sec:cost_analysis}), making it a practical plug-and-play framework for various diffusion backbones.

\subsection{Intra Contrastive Loss (IntraCL) for LF}
\label{sec:intracl}

As observed in \refsec{sec:observations}~(1, 2) and visualized in \refig{fig:in_batch_sims}, LF feature maps are highly shared across training samples, whereas HF feature maps are instance-specific. Consequently, inter-image negatives for LF are unreliable and using in-batch negatives for LF easily introduces false negatives because many images share similar coarse structures. We therefore introduce an LF-oriented IntraCL that compares a student only against its teacher and a randomly sampled layer within the same network. This stabilizes globally shared structures and promotes layer-wise refinement of LF representations.

Concretely, let $\mathbf{F}^{(i)}$ denote the feature map from an intermediate layer $i$ of the diffusion backbone (U-Net or DiT). We obtain the LF representation $\mathbf{F}^{(i)}_{\text{LF}}$ by frequency-decomposing $\mathbf{F}^{(i)}$ via 2D FFT and applying a predefined LF mask to its magnitude spectrum. Let $\mathbf{F}^{(n)}_{\text{LF}}$ denote the LF representation of the final-layer feature map (the teacher), and let $\mathbf{F}^{(j)}_{\text{LF}}$ be the LF representation of a randomly sampled student layer $j$.

We define the similarity function $\mathrm{sim}(\cdot,\cdot)$ as the cosine similarity. This is computed as the dot product between L2-normalized feature maps $\hat{\mathbf{F}} = \mathbf{F} / \|\mathbf{F}\|_2$. 
The similarity scores are \( s^{(i)}_{+,\text{LF}}=\mathrm{sim}(\hat{\mathbf{F}}^{(i)}_{\text{LF}},\hat{\mathbf{F}}^{(n)}_{\text{LF}})\;;\;
s^{(i)}_{-,\text{LF}}=\mathrm{sim}(\hat{\mathbf{F}}^{(i)}_{\text{LF}},\hat{\mathbf{F}}^{(j)}_{\text{LF}}) \),
where \( j \) is sampled uniformly at each step from \( \mathcal{J}_i=\{1,\dots,n\}\setminus\{i,n\} \), \( j\sim\mathrm{Unif}(\mathcal{J}_i) \).

The IntraCL is formulated as a log-softmax over these scores:
\begin{equation}
\mathcal{L}^{(i)}_{\text{IntraCL}}
= - \log
\frac{
\exp\big(s^{(i)}_{+,\text{LF}}\big)
}{
\exp\big(s^{(i)}_{+,\text{LF}}\big) + \exp\big(s^{(i)}_{-,\text{LF}}\big)
}.
\label{eq:intra}
\end{equation}
This self-distillation objective pulls the student toward the teacher's LF representation. Because the teacher and students operate within the same high-dimensional feature space, this approach avoids the domain mismatch problem and provides a consistent and domain-compatible distillation signal to stabilize the network's understanding of global structures.

\subsection{Inter Contrastive Loss (InterCL) for HF}
\label{sec:intercl}

\begin{figure}
    \centering
    \includegraphics[width=0.7\linewidth]{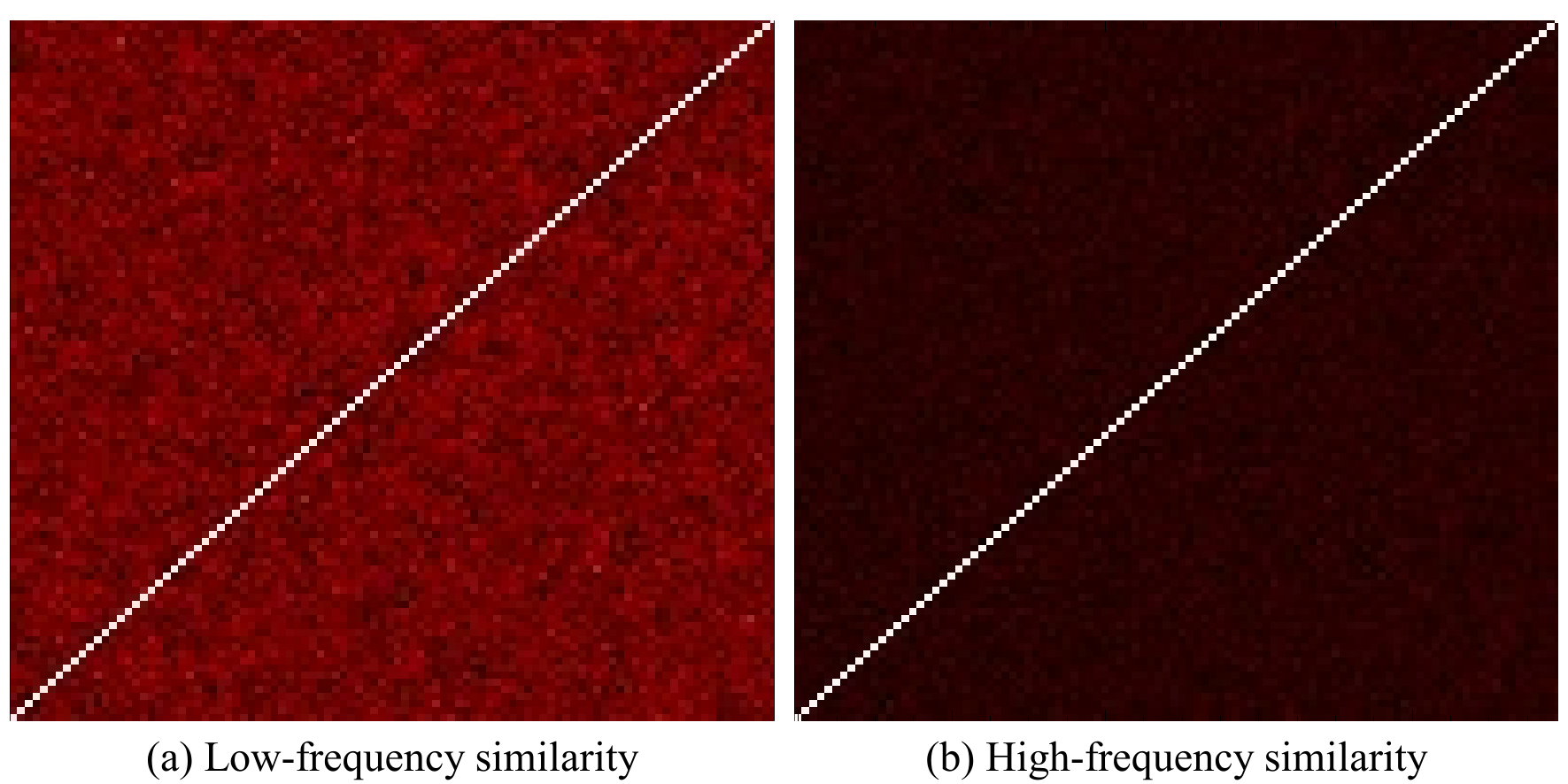}
  \caption{
    \textbf{Visualization of feature maps similarity matrices across training samples in different frequencies }
    (brighter/redder indicates higher similarity). 
    (a) LF exhibits strong cross sample similarity, 
    reflecting shared structural information and motivating the use of \textbf{IntraCL}~(\refsec{sec:intracl}) for stabilizing global structure learning. 
    (b) HF shows weak cross sample similarity and strong sample specific variation, 
    justifying the use of \textbf{InterCL}~(\refsec{sec:intercl}) to promote fine grained, instance level discrimination without over sharpening. Detailed descriptions of the training samples are in \refsec{sec:settings}
  }
  \label{fig:in_batch_sims}
\end{figure}

As observed in \refsec{sec:observations}~(1, 2) and confirmed by the cross-sample statistics in \refig{fig:in_batch_sims}, HF components encode fine-grained, sample-specific details and show low cross-sample similarity, which makes in-batch negatives informative and valid. Accordingly, we introduce an HF-oriented InterCL that attracts a student’s HF feature map to the teacher’s HF feature map while repelling it from two types of negatives: (i) a random-layer HF feature map within the same image (to enforce layer-wise progression) and (ii) \textit{in-batch} HF feature maps from other images (to encourage instance discrimination). This objective directly counteracts the LF bias and delivers targeted optimization signals to the slowly converging HF components, sharpening image-specific details.

Analogous to the LF branch, we obtain the HF representation $\mathbf{F}^{(i)}_{\text{HF}}$ by applying the predefined HF mask to the 2D FFT magnitude spectrum of the intermediate-layer feature map $\mathbf{F}^{(i)}$. Using the L2-normalized feature maps $\hat{\mathbf{F}}_{\text{HF}}$, we define the positive and random-layer negative similarity scores: \( s^{(i)}_{+,\text{HF}}=\mathrm{sim}(\hat{\mathbf{F}}^{(i)}_{\text{HF}},\hat{\mathbf{F}}^{(n)}_{\text{HF}})\;;\;
s^{(i)}_{-,\text{HF}}=\mathrm{sim}(\hat{\mathbf{F}}^{(i)}_{\text{HF}},\hat{\mathbf{F}}^{(j)}_{\text{HF}}) \), where $j$ is sampled uniformly from $\mathcal{J}_i=\{1,\dots,n\}\setminus\{i,n\}$.

Let $x^-$ denote the L2-normalized HF feature maps from other training samples in the current batch. We aggregate these in-batch negatives as: \(
S^{(i)}_{\text{neg}}=\sum_{x^-}\exp\!\big(\mathrm{sim}(\hat{\mathbf{F}}^{(i)}_{\text{HF}},x^-)\big).
\)

The InterCL loss is then formulated as a log-softmax over the positive, the random-layer negative, and the \textit{in-batch} negatives. The HF branch and its InterCL objective are depicted in \refig{fig:main}(c).
\begin{equation}
\mathcal{L}^{(i)}_{\text{InterCL}}
= - \log
\frac{
\exp\big(s^{(i)}_{+,\text{HF}}\big)
}{
\exp\big(s^{(i)}_{+,\text{HF}}\big) +
\exp\big(s^{(i)}_{-,\text{HF}}\big) +
S^{(i)}_{\text{neg}}
}.
\label{eq:inter}
\end{equation}

Here, the random-layer negative $j$ enforces layer-wise progression, while the in-batch term $S^{(i)}_{\text{neg}}$ encourages instance-aware discrimination. This objective directly counteracts the LF bias and provides targeted optimization signals to the slowly converging HF components, as shown in \refig{fig:layer-wise-cos}, helping the model reconstruct fine, HF details.

\subsection{Frequency-based Adaptive Weight (FAW)}
\label{sec:faw}

As observed in \refsec{sec:observations}~(3) and \refig{fig:layer-wise-cos}, diffusion backbones follow a depth-wise ``low-first, high-later'' frequency hierarchy: early layers prioritize stabilizing LF representations, whereas HF similarity emerges chiefly in later layers. To align the self-distillation losses (IntraCL, InterCL) with this hierarchy, we introduce Frequency-based Adaptive Weight (FAW), which decomposes self-distillation across layer depth and frequency bands and, at each layer, sets the loss weight for each band according to its relative difference to the final layer. This yields normalized per-layer LF/HF weights that directly scale the corresponding distillation losses.
The effectiveness of this FAW mechanism is empirically validated in \refsec{sec:a4}

We use the same predefined binary masks $M_{\text{LF}}$ and $M_{\text{HF}}$ from Sections~\ref{sec:intracl} and \ref{sec:intercl}. Let $\big|\mathbf{F}^{(i)}\big|$ be the FFT magnitude at layer $i$. The per-frequency average magnitudes are:
\begin{equation}
\resizebox{0.8\columnwidth}{!}{$
E^{(i)}_{\text{LF}}=\frac{\sum \left(\big|\mathbf{F}^{(i)}\big|\odot M_{\text{LF}}\right)}{\sum M_{\text{LF}}+\varepsilon}
\;,\quad
E^{(i)}_{\text{HF}}=\frac{\sum \left(\big|\mathbf{F}^{(i)}\big|\odot M_{\text{HF}}\right)}{\sum M_{\text{HF}}+\varepsilon}.
$}
\label{eq:faw_energy}
\end{equation}

Here, \(n\) denotes the index of the final (teacher) layer and \(\varepsilon\) a small constant (e.g., \(10^{-9}\)) for numerical stability; we then calculate the relative difference to the final layer \(n\):

\begin{equation}
\resizebox{0.8\columnwidth}{!}{$
\Delta^{(i)}_{\text{LF}} = 
\frac{\big|E^{(n)}_{\text{LF}} - E^{(i)}_{\text{LF}}\big|}{E^{(i)}_{\text{LF}} + \varepsilon},\quad
\Delta^{(i)}_{\text{HF}} =
\frac{\big|E^{(n)}_{\text{HF}} - E^{(i)}_{\text{HF}}\big|}{E^{(i)}_{\text{HF}} + \varepsilon}.
$}
\label{eq:faw_delta}
\end{equation}
Finally, the weights $\mathbf{w}^{(i)}$ are computed using an inverse discrepancy formulation, mitigating scale-induced spectral bias.

\begin{equation}
\resizebox{0.8\columnwidth}{!}{$
\mathbf{w}^{(i)}_{\text{LF}} = \frac{1}{1 + \Delta^{(i)}_{\text{LF}}},
\quad
\mathbf{w}^{(i)}_{\text{HF}} = \frac{1}{1 + \Delta^{(i)}_{\text{HF}}},
$}
\label{eq:faw_softmax}
\end{equation}

Given the normalized, per-layer weights \(\mathbf{w}^{(i)}\) in Eq.~\ref{eq:faw_softmax}, we combine the LF/HF contrastive objectives into a single FAW-only self-distillation term by linearly weighting the branches as follows:

\begin{equation}
\mathcal{L}^{(i)}_{\text{FRAMER}}
= w^{(i)}_{\text{LF}}\,\mathcal{L}^{(i)}_{\text{IntraCL}}
+ w^{(i)}_{\text{HF}}\,\mathcal{L}^{(i)}_{\text{InterCL}}.
\label{eq:framer_faw}
\end{equation}

\subsection{Frequency-based Alignment Modulation (FAM)}
\label{sec:fam}

As observed in \refsec{sec:observations}~(4) and \refig{fig:layer-wise-cos}, the student–teacher alignment increases along layer depth: in early layers, student features are far from the final-layer teacher, so self-distillation from earlier layers can destabilize optimization, whereas those from later layers are more helpful. We therefore propose Frequency-based Alignment Modulation (FAM), which gates the self-distillation strength by the current LF/HF alignment, suppressing it when alignment strength is low and adaptively increasing as a degree of alignment improves.
The effectiveness of this FAM mechanism is empirically validated in \refsec{sec:a4}.

We first define the alignment scores $a^{(i)}$ as the cosine similarity between the L2-normalized student and teacher feature maps (using $\hat{\mathbf{F}}$ defined in \refsec{sec:intracl}), passed through a ReLU to ensure non-negativity:
\begin{equation}
\resizebox{0.8\columnwidth}{!}{$
a^{(i)}_{\text{LF}} = \mathrm{ReLU}\!\big(\mathrm{sim}(\hat{\mathbf{F}}^{(i)}_{\text{LF}}, \hat{\mathbf{F}}^{(n)}_{\text{LF}})\big),\quad
a^{(i)}_{\text{HF}} = \mathrm{ReLU}\!\big(\mathrm{sim}(\hat{\mathbf{F}}^{(i)}_{\text{HF}}, \hat{\mathbf{F}}^{(n)}_{\text{HF}})\big).
$}
\label{eq:fam_align}
\end{equation}
We then use these alignment scores, with detached gradients, to gate the FAW weights:
\begin{equation}
\resizebox{0.8\columnwidth}{!}{$
\tilde{w}^{(i)}_{\text{LF}} = w^{(i)}_{\text{LF}} \cdot \mathrm{stopgrad}\!\big(a^{(i)}_{\text{LF}}\big),\quad
\tilde{w}^{(i)}_{\text{HF}} = w^{(i)}_{\text{HF}} \cdot \mathrm{stopgrad}\!\big(a^{(i)}_{\text{HF}}\big).
$}
\label{eq:fam_gate}
\end{equation}
The final, layer-wise distillation loss for our FRAMER framework is the weighted sum of the two contrastive losses:
{\small
\begin{equation}
\mathcal{L}^{(i)}_{\text{FRAMER}}
= \tilde{w}^{(i)}_{\text{LF}}\;\mathcal{L}^{(i)}_{\text{IntraCL}}
+ \tilde{w}^{(i)}_{\text{HF}}\;\mathcal{L}^{(i)}_{\text{InterCL}}.
\label{eq:fam_final}
\end{equation}
}
\begin{table*}[t]
\scriptsize
\renewcommand{\arraystretch}{1.15}
\setlength{\tabcolsep}{3pt}
\centering
\caption{\textbf{Quantitative comparison of real-world image super-resolution methods.} We evaluate both fidelity metrics (PSNR$\uparrow$, SSIM$\uparrow$, LPIPS$\downarrow$) and perceptual quality metrics (NIQE$\downarrow$, MANIQA$\uparrow$, MUSIQ$\uparrow$).
Methods are grouped by architecture type (Swin-based, U-Net-based, DiT-based).
\rc{\textbf{Best}} and \bc{\underline{Second best}} results are highlighted. The \gc{green} percentages for our FRAMER models indicate the relative improvement over their respective baselines, PiSA-SR (for FRAMER$_{\mathrm{U}}$) and DiT4SR (for FRAMER$_{\mathrm{D}}$).}
\scalebox{0.7}{
\begin{tabularx}{\textwidth}{l l *{2}{Y}| *{3}{Y} *{3}{Y}}
\toprule
\multicolumn{2}{c}{} &
\multicolumn{2}{c}{\textbf{Swin-based}} &
\multicolumn{3}{c}{\textbf{U-Net-based}} &
\multicolumn{3}{c}{\textbf{DiT-based}} \\
\cmidrule(lr){3-4}\cmidrule(lr){5-7}\cmidrule(lr){8-10}
\textbf{Datasets} & \textbf{Metrics} &
\makecell{\scriptsize SwinIR~\cite{liang2021swinir}\\\scriptsize(ICCV'21)} &
\makecell{\scriptsize ResShift~\cite{yue2023resshift}\\\scriptsize(NeurIPS'23)} &
\makecell{\scriptsize SeeSR~\cite{wu2024seesr}\\\scriptsize(CVPR'24)} &
\makecell{\scriptsize PiSA-SR~\cite{sun2025pixel}\\\scriptsize(CVPR'25)} &
\makecell{\scriptsize FRAMER$_{\mathrm{U}}$\\\scriptsize(\textbf{Ours})} &
\makecell{\scriptsize DreamClear~\cite{ai2024dreamclear}\\\scriptsize(NeurIPS'24)} &
\makecell{\scriptsize DiT4SR~\cite{duan2025dit4sr}\\\scriptsize(ICCV'25)} &
\makecell{\scriptsize FRAMER$_{\mathrm{D}}$\\\scriptsize(\textbf{Ours})} \\
\midrule
\textbf{DrealSR} & PSNR $\uparrow$   & 25.51 & 25.97 & 25.92 & \bc{\underline{26.18}} & \rc{\textbf{26.96}} (\gc{+3.0\%}) & 23.41 & 23.64 & 24.73 (\gc{+4.6\%}) \\
                 & SSIM $\uparrow$   & 0.742 & 0.704 & 0.734 & \bc{\underline{0.752}} & \rc{\textbf{0.786}} (\gc{+4.5\%}) & 0.652 & 0.640 & 0.687 (\gc{+7.3\%}) \\
                 & LPIPS $\downarrow$& \bc{\underline{0.366}} & 0.520 & 0.395 & 0.368 & \rc{\textbf{0.333}} (\gc{+9.5\%}) & 0.493 & 0.442 & 0.412 (\gc{+6.8\%}) \\
                 & NIQE $\downarrow$ & 6.555 & 8.763 & 6.452 & 6.136 & \rc{\textbf{5.386}} (\gc{+12.2\%}) & 6.481 & 6.780 & \bc{\underline{5.959}} (\gc{+12.1\%}) \\
                 & MANIQA $\uparrow$ & 0.330 & 0.341 & 0.507 & 0.490 & \rc{\textbf{0.595}} (\gc{+21.4\%}) & 0.418 & 0.441 & \bc{\underline{0.514}} (\gc{+16.6\%}) \\
                 & MUSIQ $\uparrow$  & 55.26 & 56.17 & 64.99 & 66.15 & \rc{\textbf{74.53}} (\gc{+12.7\%}) & 57.68 & 64.93 & \bc{\underline{68.47}} (\gc{+5.5\%}) \\
\midrule
\textbf{RealSR}  & PSNR $\uparrow$   & 23.56 & 23.51 & 23.69 & \bc{\underline{24.02}} & \rc{\textbf{24.81}} (\gc{+3.3\%}) & 21.67 & 21.94 & 23.23 (\gc{+5.9\%}) \\
                 & SSIM $\uparrow$   & 0.719 & 0.697 & 0.700 & \bc{\underline{0.719}} & \rc{\textbf{0.746}} (\gc{+3.8\%}) & 0.698 & 0.640 & 0.679 (\gc{+6.1\%}) \\
                 & LPIPS $\downarrow$& 0.378 & 0.497 & 0.386 & \bc{\underline{0.355}} & \rc{\textbf{0.328}} (\gc{+7.6\%}) & 0.489 & 0.414 & 0.371 (\gc{+10.4\%}) \\
                 & NIQE $\downarrow$ & 5.605 & 7.174 & \bc{\underline{5.337}} & 5.902 & 5.513 (\gc{+6.6\%}) & 6.691 & 6.262 & \rc{\textbf{5.083}} (\gc{+18.8\%}) \\
                 & MANIQA $\uparrow$ & 0.359 & 0.338 & \bc{\underline{0.542}} & 0.412 & 0.484 (\gc{+17.5\%}) & 0.436 & 0.459 & \rc{\textbf{0.564}} (\gc{+22.9\%}) \\
                 & MUSIQ $\uparrow$  & 61.47 & 60.67 & 69.80 & 68.20 & \bc{\underline{70.03}} (\gc{+2.7\%}) & 64.89 & 67.67 & \rc{\textbf{72.24}} (\gc{+6.8\%}) \\
\midrule
\textbf{RealLR200}& NIQE $\downarrow$ & \bc{\underline{3.693}} & 5.305 & 3.919 & 4.410 & 4.381 (\gc{+0.7\%}) & \rc{\textbf{3.229}} & 4.342 & 4.038 (\gc{+7.0\%}) \\
                 & MANIQA $\uparrow$ & 0.343 & 0.361 & 0.472 & 0.499 & \bc{\underline{0.525}} (\gc{+5.2\%}) & 0.475 & 0.514 & \rc{\textbf{0.552}} (\gc{+7.4\%}) \\
                 & MUSIQ $\uparrow$  & 64.59 & 61.24 & 71.63 & 71.95 & \bc{\underline{73.38}} (\gc{+2.0\%}) & 67.63 & 72.29 & \rc{\textbf{74.30}} (\gc{+2.8\%}) \\
\midrule
\textbf{RealLQ250}& NIQE $\downarrow$ & \bc{\underline{3.660}} & 5.274 & 3.896 & 4.138 & 4.083 (\gc{+1.3\%}) & \rc{\textbf{3.503}} & 4.199 & 3.907 (\gc{+7.0\%}) \\
                 & MANIQA $\uparrow$ & 0.357 & 0.394 & 0.490 & 0.471 & \bc{\underline{0.516}} (\gc{+9.6\%}) & 0.427 & 0.507 & \rc{\textbf{0.546}} (\gc{+7.7\%}) \\
                 & MUSIQ $\uparrow$  & 65.09 & 62.87 & 72.19 & 71.19 & \bc{\underline{73.12}} (\gc{+2.7\%}) & 68.40 & 72.19 & \rc{\textbf{73.91}} (\gc{+2.4\%}) \\
\bottomrule
\end{tabularx}}
\label{tab:results_new}
\end{table*}

\paragraph{Total objective.}

The final training objective combines the standard noise-prediction loss ($\mathcal{L}_{\text{noise}}$) with the sum of our layer-wise, frequency-gated distillation terms:

\begin{equation}
\mathcal{L}_{\text{total}}
= \mathcal{L}_{\text{noise}}
+ \sum_{i=1}^{N} \mathcal{L}^{(i)}_{\text{FRAMER}}.
\label{eq:total}
\end{equation}

This full objective aligns self-distillation with the internal frequency hierarchy, delivering targeted frequency-aware signals that counteract LF bias~\cite{choi2022perception,falck2025fourier} and, via FAW and FAM, stabilize optimization by gating distillation to layer-wise discrepancy and alignment so that early layers are stabilized and HF details in later layers are sharpened while balancing globally coherent structure with instance-specific detail. All auxiliary heads and losses are strictly training-only and removed at test time with no architectural or inference changes, making FRAMER a plug-and-play training method that leverages the strong priors of large pre-trained T2I diffusion backbones such as Stable Diffusion 2~\cite{SD2} (U-Net~\cite{ronneberger2015u}) and SD3~\cite{SD3} (DiT~\cite{peebles2023scalable}) to improve perceptual realism and high-frequency fidelity.


\section{Experiments}

\subsection{Experimental Settings}
\label{sec:settings}
\textbf{Datasets.} To ensure fair comparisons, we follow the same experimental setup as SeeSR~\cite{wu2024seesr} and DiT4SR~\cite{duan2025dit4sr}. 
Our training set is composed of a mixture of images from DIV2K~\cite{agustsson2017ntire}, DIV8K~\cite{gu2019div8k}, Flickr2K~\cite{timofte2017ntire},
and the initial 10K face images from FFHQ~\cite{karras2019style}.
The Real-ESRGAN~\cite{wang2021real} degradation pipeline for SR training is adopted to generate LR–HR training pairs using the same parameter configuration as SeeSR. The input LR and HR resolutions are set to $128\times128$ and $512\times512$, respectively. For evaluation, we employ four real-world datasets: DrealSR~\cite{wei2020component}, RealSR~\cite{cai2019toward}, RealLR200~\cite{wu2024seesr}, and RealLQ250~\cite{ai2024dreamclear}. All experiments are conducted with an upscaling factor of ×4.

\noindent\textbf{Metrics.} To ensure comparability with prior works, we report both full-reference and no-reference metrics. Specifically, we include PSNR and SSIM~\cite{wang2004ssim} as full-reference measures for distortion/error, while acknowledging their limited alignment with human perception in restoration tasks; for perceptual fidelity we adopt LPIPS~\cite{zhang2018unreasonable}. For no-reference quality assessment, we use NIQE~\cite{mittal2012making} as an opinion-unaware metric, and MANIQA~\cite{yang2022maniqa} and MUSIQ~\cite{ke2021musiq} as recent learning-based NR-IQA measures.

\noindent\textbf{Implementation Details.}
All implementation details, including loss functions, frequency decomposition, training iterations, etc., are provided in the \textit{Suppl.} for reproducibility.

\subsection{Comparison with Other Methods}

We compare our method with state-of-the-art Real-ISR methods, including Swin Transformer-based methods, U-Net-based methods, and DiT-based methods. Among these, our FRAMER is instantiated on the U-Net-based PiSA-SR and the DiT-based DiT4SR to validate its architectural flexibility. FRAMER is trained in a fully plug-and-play manner under exactly the same settings as each corresponding baseline, without any architecture-specific tuning. Unlike DiT, whose feature resolutions remain identical across layers, U-Net changes the feature resolution between stages; therefore, we insert a 1×1 convolution and a resize operation to align dimensions when integrating FRAMER into the U-Net, as consistent feature shapes are required for self-distillation.

\noindent\textbf{Quantitative Comparisons.} We conduct quantitative evaluations of our proposed FRAMER framework against recent state-of-the-art Real-ISR methods across four real-world benchmarks, as shown in \reftbl{tab:results_new}. On the DrealSR and RealSR datasets, FRAMER achieves performance on par with or superior to advanced approaches such as SeeSR and DiffBIR, demonstrating strong fidelity and perceptual quality. On the more challenging RealLR200 and RealLQ250 datasets, our method exhibits clear superiority, ranking first across all perceptual quality metrics. These results collectively highlight the effectiveness of FRAMER in producing high-quality and visually realistic restorations, benefiting from its robust generative prior and architecture-agnostic design.

\noindent\textbf{Qualitative Comparisons.} The qualitative results compared with recent Real-ISR methods are illustrated in \refig{fig:qualitative_comparisons}.
As shown, our method produces clearer textures and more faithful structures than existing approaches, particularly under severe degradations.
Both FRAMER$_{\mathrm{U}}$ and FRAMER$_{\mathrm{D}}$ effectively recover fine details while preserving natural appearance, highlighting the strong restoration capability of our framework.
More comprehensive qualitative comparisons can be found in the \textit{Supplementary Material}.
\section{Ablation Study}

To validate the effectiveness of our proposed FRAMER, we conduct comprehensive ablations on its key components: (1) the frequency-specific contrastive distillation objective, (2) the design of IntraCL and InterCL, and (3) the adaptive mechanisms FAW and FAM. Unless otherwise noted, all experiments share the same setting: training is performed under the same conditions as \refsec{sec:settings}, and evaluation is conducted on the RealSR dataset.

\subsection{Effectiveness of the Distillation Objective}
\label{sec:a1}

\begin{table}[h]
\centering
\caption{\textbf{Ablation on the distillation objective.} Metrics are on RealSR. The \textbf{best} per column is in \textbf{bold}.}
\label{tab:ablation_objective}
\begin{adjustbox}{width=\linewidth}
\begin{tabular}{l cc c ccc}
\toprule
\multirow{2}{*}{Method} & \multicolumn{2}{c}{Distillation Setup} & \multicolumn{2}{c}{Quality Metrics} & \multicolumn{2}{c}{Perceptual Metrics} \\
\cmidrule(lr){2-3} \cmidrule(lr){4-5} \cmidrule(lr){6-7}
 & Freq. Decomp. & Loss Type & PSNR$\uparrow$ & LPIPS$\downarrow$ & MUSIQ$\uparrow$ & MANIQA$\uparrow$ \\
\midrule
\texttt{Baseline}              & $\times$ & $\times$ & 21.94 & 0.41 & 67.67 & 0.459 \\
\texttt{+MSE Distill.}         & $\times$ & MSE & 22.36 & 0.41 & 68.11 & 0.467 \\
\texttt{+MSE-Freq Distill.}    & \checkmark & MSE & 22.59 & 0.39 & 65.74 & 0.417 \\
\texttt{+CL-Freq Distill.}     & \checkmark & CL  & \textbf{23.04} & \textbf{0.38} & \textbf{69.01} & \textbf{0.533} \\
\bottomrule
\end{tabular}
\end{adjustbox}
\end{table}

As shown in \reftbl{tab:ablation_objective}, we first test the hypothesis that a frequency-specific contrastive objective is preferable to L2-based self-distillation for Real-ISR. Adding frequency-agnostic L2 between the teacher and students (``{\texttt{+MSE Distill.}}'') yields modest gains over ``{\texttt{Baseline}}'' but does not correct the LF bias of the standard noise-prediction loss~\cite{ho2020denoising}. Introducing frequency decomposition with L2 (``{\texttt{+MSE-Freq Distill.}}'') improves distortion metrics yet harms perceptual metrics, consistent with domain mismatch when regressing high-dimensional features to pixel/Fourier targets~\cite{tian2020contrastive,zhao2022decoupled}. In contrast, our frequency-specific contrastive variant (``{\texttt{+CL-Freq Distill.}}'') achieves the best results across all metrics by providing a domain-compatible signal in the same feature space~\cite{SDSSL,SSSD,zhang2019your} as the backbone’s representations, while explicitly counteracting the LF dominance with frequency-aware distillation from the final-layer teacher.

\subsection{Analysis of Contrastive Learning Components}
\label{sec:a2}

The detailed results for this analysis are available in \reftbl{tab:ablation_cl_components}. \textbf{Teacher selection.} When fixed to ``{\texttt{Negative: Random Layer}}'', using a ``{\texttt{Random Layer}}'' as the teacher underperforms, indicating that arbitrary references do not consistently reflect the target representation to be learned. Moving the teacher to ``{\texttt{Final Layer -1}}'' or ``{\texttt{Final Layer -2}}'' further weakens either distortion or perceptual quality, aligning with the layer-depth-wise frequency progression: early layers stabilize LF, while the true ``{\texttt{Final Layer}}'' best captures converged HF details. Thus, the ``{\texttt{Final Layer}}'' teacher provides the most informative target for both IntraCL and InterCL. \textbf{Negative selection.} Fixing ``{\texttt{Teacher: Final Layer}}'', drawing the negative from the ``{\texttt{Previous Layer}}'' leads to weaker signals due to high correlation with the student. Sampling a ``{\texttt{Random Layer}}'' negative enforces comparisons across layer-depth, strengthens the margin, and—combined with the in-batch negatives in InterCL yields a more diverse and informative set of contrasts. The combination of ``{\texttt{Final Layer}}'' teacher and ``{\texttt{Random Layer}}'' negative achieves the strongest overall performance.

\begin{table}[h]
\centering
\caption{\textbf{Ablation on contrastive components.}}
\label{tab:ablation_cl_components}
\begin{adjustbox}{width=\linewidth}
\begin{tabular}{l c ccc}
\toprule
\multirow{2}{*}{Method} & \multicolumn{2}{c}{Quality Metrics} & \multicolumn{2}{c}{Perceptual Metrics} \\
\cmidrule(lr){2-3} \cmidrule(lr){4-5}
 & PSNR$\uparrow$ & LPIPS$\downarrow$ & MUSIQ$\uparrow$ & MANIQA$\uparrow$ \\
\midrule
\multicolumn{5}{l}{Part 1: Teacher Selection (Negative=\texttt{Random Layer})} \\
\quad \texttt{Random Layer}     & 22.63 & 0.39 & 67.23 & 0.469 \\
\quad \texttt{Final Layer -1}    & 22.15 & 0.40 & 67.45 & 0.476 \\
\quad \texttt{Final Layer -2}    & 22.62 & 0.39 & 65.82 & 0.438 \\
\midrule
\multicolumn{5}{l}{Part 2: Negative Selection (Teacher=\texttt{Final Layer})} \\
\quad \texttt{Previous Layer}  & 22.23 & 0.39 & 66.44 & 0.445 \\
\midrule
\textbf{\texttt{Ours: Final Teacher + Random Neg.}} & \textbf{23.04} & \textbf{0.38} & \textbf{69.01} & \textbf{0.533} \\
\bottomrule
\end{tabular}
\end{adjustbox}
\end{table}

\subsection{Impact of Frequency-Specific Design}
\label{sec:a3}

\reftbl{tab:performance} compares the performance of different loss combinations for the LF and HF bands.
\textbf{Individual frequency branches.} Applying only IntraCL on LF (``{\texttt{A}}'') stabilizes globally shared structure but yields suboptimal improvement in HF details, giving limited perceptual gains. Applying only InterCL on HF (``{\texttt{D}}'') sharpens instance-specific details and improves perceptual metrics, but the lack of LF stabilization can slightly hurt overall fidelity. Symmetrically, placing a single loss on the wrong band (``{\texttt{B}}'' or ``{\texttt{C}}'') underutilizes the frequency-progressive hierarchy.
\textbf{LF/HF combinations:} Pairing IntraCL for LF with InterCL for HF (``{\texttt{H}}'') aligns distillation strength with the low-first, high-later hierarchy and yields the best distortion–perception trade-off. Reversing the assignment (``{\texttt{E}}'') or using the same contrast on both bands (``{\texttt{F}}'', ``{\texttt{G}}'') degrades either perceptual realism or fidelity, confirming that LF benefits from pairwise stabilization while HF benefits from instance discrimination with in-batch negatives.

\begin{table}[h]
\centering
\caption{\textbf{Ablation on the combination of contrastive losses for LF/HF bands.}}
\begin{adjustbox}{width=\linewidth}
\begin{tabular}{c c c c c c c}
\toprule
\multirow{2}{*}{Variant} & \multicolumn{2}{c}{Loss Options} & \multicolumn{2}{c}{Quality Metrics} & \multicolumn{2}{c}{Perceptual Metrics} \\
\cmidrule(lr){2-3} \cmidrule(lr){4-5} \cmidrule(lr){6-7}
 & Low Freq. & High Freq. & PSNR$\uparrow$ & LPIPS$\downarrow$ & MUSIQ$\uparrow$ & MANIQA$\uparrow$ \\
\midrule
\texttt{Baseline} & $\times$ & $\times$ & 21.94 & 0.41 & 67.67 & 0.459 \\ \hline \hline
\texttt{A} & IntraCL & $\times$ & 22.86 & 0.38 & 65.92 & 0.466 \\
\texttt{B} & InterCL & $\times$ & 22.62 & 0.41 & 66.22 & 0.479 \\
\texttt{C} & $\times$ & IntraCL & 22.27 & 0.39 & 67.98 & 0.509 \\ 
\texttt{D} & $\times$ & InterCL & 22.47 & 0.39 & 68.70 & 0.513 \\
\hline
\texttt{E}    & InterCL & IntraCL & 22.21 & 0.41 & 67.56 & 0.484 \\
\texttt{F}    & InterCL & InterCL & 21.90 & 0.41 & 67.21 & 0.465 \\
\texttt{G}    & IntraCL & IntraCL   & 22.85 & 0.39 & 65.25 & 0.435 \\
\texttt{H (Ours)}    & IntraCL & InterCL   & \textbf{23.04} & \textbf{0.38} & \textbf{69.01} & \textbf{0.533} \\
\bottomrule
\end{tabular}
\end{adjustbox}
\label{tab:performance}
\end{table}

\subsection{Effectiveness of Adaptive Mechanisms}
\label{sec:a4}

We validate the effectiveness of FAW and FAM in \reftbl{tab:fawfam}.
\textbf{Adaptive weighting with FAW:} ``{\texttt{FAW}}'' computes normalized per-layer weights from student–teacher frequency discrepancies, aligning loss strength with the frequency-progressive hierarchy. Even without ``{\texttt{FAM}}'', ``{\texttt{FAW}}'' improves perceptual quality while maintaining distortion metrics.
\textbf{Alignment gating with FAM:} ``{\texttt{FAM}}'' gates losses by the current LF/HF alignment degree to the final-layer teacher, reducing distillation strength when alignment is low (early layers) and increasing it as alignment improves (later layers). This stabilizes optimization and slightly improves both distortion and perception.
\textbf{Combined effect:} Using both ``{\texttt{FAW}}'' and ``{\texttt{FAM}}'' yields the strongest overall performance, indicating that discrepancy-aware weighting and alignment-aware gating are complementary: ``{\texttt{FAW}}'' allocates distillation strength across frequencies and layer-depth, while ``{\texttt{FAM}}'' schedules its intensity according to layer-wise alignment. Together they best counter the LF bias and sharpen HF details without destabilizing early layers.

\begin{table}[h]
\centering
\caption{\textbf{Ablation study on the effectiveness of FAW and FAM.}}
\begin{adjustbox}{width=\linewidth}
\begin{tabular}{c c c c c c c}
\toprule
\multirow{2}{*}{Variant} & \multicolumn{2}{c}{Adaptive Distillation} & \multicolumn{2}{c}{Quality Metrics} & \multicolumn{2}{c}{Perceptual Metrics} \\
\cmidrule(lr){2-3} \cmidrule(lr){4-5} \cmidrule(lr){6-7}
 & FAW & FAM & PSNR$\uparrow$ & LPIPS$\downarrow$ & MUSIQ$\uparrow$ & MANIQA$\uparrow$ \\
\midrule
\texttt{Baseline} & $\times$ & $\times$ & 21.94 & 0.41 & 67.67 & 0.459 \\ \hline
\texttt{CL only}  & $\times$ & $\times$ & 23.04 & 0.38 & 69.01 & 0.533 \\
\texttt{FAW only} & \checkmark & $\times$ & 22.93 & 0.38 & 70.50 & 0.552 \\
\texttt{FAM only} & $\times$ & \checkmark & 23.06 & 0.37 & 69.43 & 0.540 \\
\texttt{FAW + FAM} & \checkmark & \checkmark & \textbf{23.23} & \textbf{0.37} & \textbf{72.24} & \textbf{0.564} \\
\bottomrule
\end{tabular}
\end{adjustbox}
\label{tab:fawfam}
\end{table}
\section{Conclusion}
FRAMER is a frequency-aware, layer-adaptive self-distillation framework for Real-ISR. It decomposes features with simple FFT masks and couples IntraCL and InterCL with FAW and FAM to deliver targeted optimization, prevent early-layer collapse, speed convergence, and enhance perceptual quality, without altering the backbone or inference. Across U-Net and DiT backbones and standard real-world benchmarks, FRAMER achieves consistent gains on both distortion and no-reference metrics, with ablations supporting the use of the final-layer feature map as teacher and random-layer plus in-batch negatives for HF.


\clearpage

\section*{Acknowledgements}
This work was supported by the National Research Foundation of Korea (NRF) grant funded by the Korea government (MSIT) (No. RS-2025-23524035). This work was partly supported by the Institute of Information \& Communications Technology Planning \& Evaluation (IITP) grant funded by the Korea government (MSIT) (No. RS-2025-25422680, Metacognitive AGI Framework and its Applications, 50\%).

{
    \small
    \bibliographystyle{ieeenat_fullname}
    \bibliography{main}

\begin{thebibliography}{53}
\providecommand{\natexlab}[1]{#1}
\providecommand{\url}[1]{\texttt{#1}}
\expandafter\ifx\csname urlstyle\endcsname\relax
  \providecommand{\doi}[1]{doi: #1}\else
  \providecommand{\doi}{doi: \begingroup \urlstyle{rm}\Url}\fi

\bibitem[Agustsson and Timofte(2017)]{agustsson2017ntire}
Eirikur Agustsson and Radu Timofte.
\newblock Ntire 2017 challenge on single image super-resolution: Dataset and study.
\newblock 2017.

\bibitem[Ai et~al.(2024)Ai, Zhou, Huang, Han, Chen, You, and Yang]{ai2024dreamclear}
Yuang Ai, Xiaoqiang Zhou, Huaibo Huang, Xiaotian Han, Zhengyu Chen, Quanzeng You, and Hongxia Yang.
\newblock Dreamclear: High-capacity real-world image restoration with privacy-safe dataset curation.
\newblock \emph{Advances in Neural Information Processing Systems}, 37:\penalty0 55443--55469, 2024.

\bibitem[Berrada et~al.(2025)Berrada, Astolfi, Hall, Havasi, Benchetrit, Romero-Soriano, Alahari, Drozdzal, and Verbeek]{berrada2025boosting}
Tariq Berrada, Pietro Astolfi, Melissa Hall, Marton Havasi, Yohann Benchetrit, Adriana Romero-Soriano, Karteek Alahari, Michal Drozdzal, and Jakob Verbeek.
\newblock Boosting latent diffusion with perceptual objectives.
\newblock In \emph{The Thirteenth International Conference on Learning Representations}, 2025.

\bibitem[Cai et~al.(2019)Cai, Zeng, Yong, Cao, and Zhang]{cai2019toward}
Jianrui Cai, Hui Zeng, Hongwei Yong, Zisheng Cao, and Lei Zhang.
\newblock Toward real-world single image super-resolution: A new benchmark and a new model.
\newblock 2019.

\bibitem[Chen and Chu(2023)]{SSSD}
Wei-Chi Chen and Wei-Ta Chu.
\newblock Sssd: Self-supervised self distillation.
\newblock In \emph{2023 IEEE/CVF Winter Conference on Applications of Computer Vision (WACV)}, pages 2769--2776, 2023.

\bibitem[Cheng et~al.(2025)Cheng, Yu, Tu, He, Chen, Guo, Zhu, Wang, Gao, and Hu]{cheng2025effective}
Kun Cheng, Lei Yu, Zhijun Tu, Xiao He, Liyu Chen, Yong Guo, Mingrui Zhu, Nannan Wang, Xinbo Gao, and Jie Hu.
\newblock Effective diffusion transformer architecture for image super-resolution.
\newblock In \emph{Proceedings of the AAAI Conference on Artificial Intelligence}, pages 2455--2463, 2025.

\bibitem[Choi et~al.(2022)Choi, Lee, Shin, Kim, Kim, and Yoon]{choi2022perception}
Jooyoung Choi, Jungbeom Lee, Chaehun Shin, Sungwon Kim, Hyunwoo Kim, and Sungroh Yoon.
\newblock Perception prioritized training of diffusion models.
\newblock In \emph{Proceedings of the IEEE/CVF conference on computer vision and pattern recognition}, pages 11472--11481, 2022.

\bibitem[Dhariwal and Nichol(2021)]{dhariwal2021diffusion}
Prafulla Dhariwal and Alexander Nichol.
\newblock Diffusion models beat gans on image synthesis.
\newblock \emph{Advances in neural information processing systems}, 34:\penalty0 8780--8794, 2021.

\bibitem[Dong et~al.(2014)Dong, Loy, He, and Tang]{dong2014learning}
Chao Dong, Chen~Change Loy, Kaiming He, and Xiaoou Tang.
\newblock Learning a deep convolutional network for image super-resolution.
\newblock In \emph{European conference on computer vision}, pages 184--199. Springer, 2014.

\bibitem[Duan et~al.(2025)Duan, Zhang, Jin, Zhang, Xiong, Zou, Ren, Guo, and Li]{duan2025dit4sr}
Zheng-Peng Duan, Jiawei Zhang, Xin Jin, Ziheng Zhang, Zheng Xiong, Dongqing Zou, Jimmy Ren, Chun-Le Guo, and Chongyi Li.
\newblock Dit4sr: Taming diffusion transformer for real-world image super-resolution.
\newblock In \emph{ICCV 2025 Poster}, 2025.
\newblock Exhibit Hall I \#1755, Poster ID 534, Oct 22, 5:45–7:45 p.m. PDT.

\bibitem[Esser et~al.(2024)Esser, Kulal, Blattmann, Entezari, M{\"u}ller, Saini, Levi, Lorenz, Sauer, Boesel, et~al.]{SD3}
Patrick Esser, Sumith Kulal, Andreas Blattmann, Rahim Entezari, Jonas M{\"u}ller, Harry Saini, Yam Levi, Dominik Lorenz, Axel Sauer, Frederic Boesel, et~al.
\newblock Scaling rectified flow transformers for high-resolution image synthesis.
\newblock In \emph{Forty-first international conference on machine learning}, 2024.

\bibitem[Falck et~al.(2025)Falck, Pandeva, Zahirnia, Lawrence, Turner, Meeds, Zazo, and Karmalkar]{falck2025fourier}
Fabian Falck, Teodora Pandeva, Kiarash Zahirnia, Rachel Lawrence, Richard Turner, Edward Meeds, Javier Zazo, and Sushrut Karmalkar.
\newblock A fourier space perspective on diffusion models.
\newblock \emph{arXiv preprint arXiv:2505.11278}, 2025.

\bibitem[Gendy et~al.(2025)Gendy, He, and Sabor]{10.1016/j.neucom.2024.128911}
Garas Gendy, Guanghui He, and Nabil Sabor.
\newblock Diffusion models for image super-resolution: State-of-the-art and future directions.
\newblock \emph{Neurocomput.}, 617\penalty0 (C), 2025.

\bibitem[Gu et~al.(2019)Gu, Lugmayr, Danelljan, Fritsche, Lamour, and Timofte]{gu2019div8k}
Shuhang Gu, Andreas Lugmayr, Martin Danelljan, Manuel Fritsche, Julien Lamour, and Radu Timofte.
\newblock Div8k: Diverse 8k resolution image dataset.
\newblock 2019.

\bibitem[Hinton et~al.(2015)Hinton, Vinyals, and Dean]{hinton2015distilling}
Geoffrey Hinton, Oriol Vinyals, and Jeff Dean.
\newblock Distilling the knowledge in a neural network.
\newblock \emph{arXiv preprint arXiv:1503.02531}, 2015.

\bibitem[Ho et~al.(2020)Ho, Jain, and Abbeel]{ho2020denoising}
Jonathan Ho, Ajay Jain, and Pieter Abbeel.
\newblock Denoising diffusion probabilistic models.
\newblock \emph{Advances in neural information processing systems}, 33:\penalty0 6840--6851, 2020.

\bibitem[Jang et~al.(2023)Jang, Kim, Yoo, Kong, Kim, and Kwak]{SDSSL}
Jiho Jang, Seonhoon Kim, Kiyoon Yoo, Chaerin Kong, Jangho Kim, and Nojun Kwak.
\newblock Self-distilled self-supervised representation learning.
\newblock In \emph{2023 IEEE/CVF Winter Conference on Applications of Computer Vision (WACV)}, pages 2828--2838, 2023.

\bibitem[Jiang et~al.(2025)Jiang, Wang, Li, Zhang, Wang, Wei, Dai, Zhang, and Wang]{jiang2025no}
Dengyang Jiang, Mengmeng Wang, Liuzhuozheng Li, Lei Zhang, Haoyu Wang, Wei Wei, Guang Dai, Yanning Zhang, and Jingdong Wang.
\newblock No other representation component is needed: Diffusion transformers can provide representation guidance by themselves.
\newblock \emph{arXiv preprint arXiv:2505.02831}, 2025.

\bibitem[Jiralerspong et~al.(2025)Jiralerspong, Earnshaw, Hartford, Bengio, and Scimeca]{jiralerspong2025shaping}
Thomas Jiralerspong, Berton Earnshaw, Jason Hartford, Yoshua Bengio, and Luca Scimeca.
\newblock Shaping inductive bias in diffusion models through frequency-based noise control.
\newblock In \emph{ICLR 2025 Workshop on Deep Generative Model in Machine Learning: Theory, Principle and Efficacy}, 2025.

\bibitem[Karras(2019)]{karras2019style}
Tero Karras.
\newblock A style-based generator architecture for generative adversarial networks.
\newblock \emph{arXiv preprint arXiv:1812.04948}, 2019.

\bibitem[Ke et~al.(2021)Ke, Wang, Wang, Milanfar, and Yang]{ke2021musiq}
Junjie Ke, Qifei Wang, Yilin Wang, Peyman Milanfar, and Feng Yang.
\newblock Musiq: Multi-scale image quality transformer.
\newblock 2021.

\bibitem[Kuznedelev et~al.(2024)Kuznedelev, Startsev, Shlenskii, and Kastryulin]{kuznedelev2024does}
Denis Kuznedelev, Valerii Startsev, Daniil Shlenskii, and Sergey Kastryulin.
\newblock Does diffusion beat gan in image super resolution?
\newblock \emph{arXiv preprint arXiv:2405.17261}, 2024.

\bibitem[Li et~al.(2025)Li, Zhao, Zhou, Xu, Hu, Chen, and Wang]{li2025fedsr}
Yueying Li, Hanbin Zhao, Jiaqing Zhou, Guozhi Xu, Tianlei Hu, Gang Chen, and Haobo Wang.
\newblock Fed{SR}: Frequency-aware enhancement for diffusion-based image super-resolution, 2025.

\bibitem[Liang et~al.(2021)Liang, Cao, Sun, Zhang, Van~Gool, and Timofte]{liang2021swinir}
Jingyun Liang, Jiezhang Cao, Guolei Sun, Kai Zhang, Luc Van~Gool, and Radu Timofte.
\newblock Swinir: Image restoration using swin transformer.
\newblock In \emph{Proceedings of the IEEE/CVF international conference on computer vision}, pages 1833--1844, 2021.

\bibitem[Lin et~al.(2024{\natexlab{a}})Lin, Zhang, Valanarasu, Wang, Gatti, and Patel]{lin2024fouriscale}
Leon Lin, Rodger Zhang, Jeya Maria~Jose Valanarasu, Haoxiang Wang, Evangelos Gatti, Prajwal~andpKalogerakis, and Vishal~M Patel.
\newblock Fouriscale: A frequency perspective on training-free high-resolution image synthesis.
\newblock In \emph{European Conference on Computer Vision (ECCV)}, 2024{\natexlab{a}}.

\bibitem[Lin et~al.(2024{\natexlab{b}})Lin, He, Chen, Lyu, Dai, Yu, Qiao, Ouyang, and Dong]{10.1007/978-3-031-73202-7_25}
Xinqi Lin, Jingwen He, Ziyan Chen, Zhaoyang Lyu, Bo Dai, Fanghua Yu, Yu Qiao, Wanli Ouyang, and Chao Dong.
\newblock Diffbir: Toward blind image restoration with generative diffusion prior.
\newblock In \emph{Computer Vision – ECCV 2024: 18th European Conference, Milan, Italy, September 29–October 4, 2024, Proceedings, Part LIX}, page 430–448, Berlin, Heidelberg, 2024{\natexlab{b}}. Springer-Verlag.

\bibitem[Liu et~al.(2023)Liu, Li, Wu, and Lee]{liu2023llava}
Haotian Liu, Chunyuan Li, Qingyang Wu, and Yong~Jae Lee.
\newblock Visual instruction tuning, 2023.

\bibitem[Ma et~al.(2025)Ma, Yu, Liu, Fang, and Wang]{Ma_2025_CVPR}
Xinyin Ma, Runpeng Yu, Songhua Liu, Gongfan Fang, and Xinchao Wang.
\newblock Diffusion model is effectively its own teacher.
\newblock In \emph{Proceedings of the IEEE/CVF Conference on Computer Vision and Pattern Recognition (CVPR)}, pages 12901--12911, 2025.

\bibitem[Medi et~al.(2025)Medi, Wang, Rampini, and Keuper]{medi2025missing}
Tejaswini Medi, Hsien-Yi Wang, Arianna Rampini, and Margret Keuper.
\newblock Missing fine details in images: Last seen in high frequencies.
\newblock \emph{arXiv e-prints}, pages arXiv--2509, 2025.

\bibitem[Mittal et~al.(2012)Mittal, Soundararajan, and Bovik]{mittal2012making}
Anish Mittal, Rajiv Soundararajan, and Alan~C Bovik.
\newblock Making a “completely blind” image quality analyzer.
\newblock \emph{IEEE Signal processing letters}, 20\penalty0 (3):\penalty0 209--212, 2012.

\bibitem[Peebles and Xie(2023)]{peebles2023scalable}
William Peebles and Saining Xie.
\newblock Scalable diffusion models with transformers.
\newblock In \emph{Proceedings of the IEEE/CVF international conference on computer vision}, pages 4195--4205, 2023.

\bibitem[Rombach et~al.(2022{\natexlab{a}})Rombach, Blattmann, Lorenz, Esser, and Ommer]{SD2}
Robin Rombach, Andreas Blattmann, Dominik Lorenz, Patrick Esser, and Bj{\"o}rn Ommer.
\newblock High-resolution image synthesis with latent diffusion models.
\newblock In \emph{Proceedings of the IEEE/CVF conference on computer vision and pattern recognition}, pages 10684--10695, 2022{\natexlab{a}}.

\bibitem[Rombach et~al.(2022{\natexlab{b}})Rombach, Blattmann, Lorenz, Esser, and Ommer]{rombach2022high}
Robin Rombach, Andreas Blattmann, Dominik Lorenz, Patrick Esser, and Bj{\"o}rn Ommer.
\newblock High-resolution image synthesis with latent diffusion models.
\newblock In \emph{Proceedings of the IEEE/CVF conference on computer vision and pattern recognition}, pages 10684--10695, 2022{\natexlab{b}}.

\bibitem[Romero et~al.(2014)Romero, Ballas, Kahou, Chassang, Gatta, and Bengio]{romero2014fitnets}
Adriana Romero, Nicolas Ballas, Samira~Ebrahimi Kahou, Antoine Chassang, Carlo Gatta, and Yoshua Bengio.
\newblock Fitnets: Hints for thin deep nets. arxiv 2014.
\newblock \emph{arXiv preprint arXiv:1412.6550}, 2014.

\bibitem[Ronneberger et~al.(2015)Ronneberger, Fischer, and Brox]{ronneberger2015u}
Olaf Ronneberger, Philipp Fischer, and Thomas Brox.
\newblock U-net: Convolutional networks for biomedical image segmentation.
\newblock In \emph{International Conference on Medical image computing and computer-assisted intervention}, pages 234--241. Springer, 2015.

\bibitem[Saharia et~al.(2022)Saharia, Ho, Chan, Salimans, Fleet, and Norouzi]{saharia2022image}
Chitwan Saharia, Jonathan Ho, William Chan, Tim Salimans, David~J Fleet, and Mohammad Norouzi.
\newblock Image super-resolution via iterative refinement.
\newblock \emph{IEEE transactions on pattern analysis and machine intelligence}, 45\penalty0 (4):\penalty0 4713--4726, 2022.

\bibitem[Sami et~al.(2024)Sami, Hasan, Dawson, and Nasrabadi]{sami2024hf}
Shoaib~Meraj Sami, Md~Mahedi Hasan, Jeremy Dawson, and Nasser Nasrabadi.
\newblock Hf-diff: High-frequency perceptual loss and distribution matching for one-step diffusion-based image super-resolution.
\newblock \emph{arXiv preprint arXiv:2411.13548}, 2024.

\bibitem[Shen et~al.(2024)Shen, Mao, and Fan]{10.1145/3696271.3696307}
Zhuoyi Shen, Maoyu Mao, and Pengfei Fan.
\newblock A primary comparison of diffusion models and generative adversarial networks for image synthesis.
\newblock In \emph{Proceedings of the 2024 7th International Conference on Machine Learning and Machine Intelligence (MLMI)}, page 225–234, New York, NY, USA, 2024. Association for Computing Machinery.

\bibitem[Sun et~al.(2025)Sun, Wu, Ma, Liu, Yi, and Zhang]{sun2025pixel}
Lingchen Sun, Rongyuan Wu, Zhiyuan Ma, Shuaizheng Liu, Qiaosi Yi, and Lei Zhang.
\newblock Pixel-level and semantic-level adjustable super-resolution: A dual-lora approach.
\newblock In \emph{Proceedings of the Computer Vision and Pattern Recognition Conference}, pages 2333--2343, 2025.

\bibitem[Tian et~al.(2020)Tian, Krishnan, and Isola]{tian2020contrastive}
Yonglong Tian, Dilip Krishnan, and Phillip Isola.
\newblock Contrastive representation distillation.
\newblock In \emph{International Conference on Learning Representations (ICLR)}, 2020.

\bibitem[Timofte et~al.(2017)Timofte, Agustsson, Van~Gool, Yang, and Zhang]{timofte2017ntire}
Radu Timofte, Eirikur Agustsson, Luc Van~Gool, Ming-Hsuan Yang, and Lei Zhang.
\newblock Ntire 2017 challenge on single image super-resolution: Methods and results.
\newblock 2017.

\bibitem[Wan et~al.(2024)Wan, Jiang, Hou, Zhang, Chen, Cheng, and Li]{wan2024controlsr}
Yuhao Wan, Peng-Tao Jiang, Qibin Hou, Hao Zhang, Jinwei Chen, Ming-Ming Cheng, and Bo Li.
\newblock Controlsr: Taming diffusion models for consistent real-world image super resolution.
\newblock \emph{arXiv preprint arXiv:2410.14279}, 2024.

\bibitem[Wang et~al.(2021)Wang, Xie, Dong, and Shan]{wang2021real}
Xintao Wang, Liangbin Xie, Chao Dong, and Ying Shan.
\newblock Real-esrgan: Training real-world blind super-resolution with pure synthetic data.
\newblock In \emph{Proceedings of the IEEE/CVF international conference on computer vision}, pages 1905--1914, 2021.

\bibitem[Wang et~al.(2024)Wang, Chai, and Chen]{wang2024frequency}
Xingjian Wang, Li Chai, and Jiming Chen.
\newblock Frequency-domain refinement with multiscale diffusion for super resolution.
\newblock \emph{arXiv preprint arXiv:2405.10014}, 2024.

\bibitem[Wang et~al.(2004)Wang, Bovik, Sheikh, and Simoncelli]{wang2004ssim}
Zhou Wang, Alan~C Bovik, Hamid~R Sheikh, and Eero~P Simoncelli.
\newblock Image quality assessment: from error visibility to structural similarity.
\newblock 2004.

\bibitem[Wei et~al.(2020)Wei, Xie, Lu, Zhan, Ye, Zuo, and Lin]{wei2020component}
Pengxu Wei, Ziwei Xie, Hannan Lu, Zongyuan Zhan, Qixiang Ye, Wangmeng Zuo, and Liang Lin.
\newblock Component divide-and-conquer for real-world image super-resolution.
\newblock 2020.

\bibitem[Woods and Bloem(2024)]{woods2024selfdistillation}
Damion Woods and Peter Bloem.
\newblock Self-distillation for diffusion models, 2024.

\bibitem[Wu et~al.(2024)Wu, Yang, Sun, Zhang, Li, and Zhang]{wu2024seesr}
Rongyuan Wu, Tao Yang, Lingchen Sun, Zhengqiang Zhang, Shuai Li, and Lei Zhang.
\newblock Seesr: Towards semantics-aware real-world image super-resolution.
\newblock In \emph{Proceedings of the IEEE/CVF conference on computer vision and pattern recognition}, pages 25456--25467, 2024.

\bibitem[Yang et~al.(2022)Yang, Wu, Shi, Lao, Gong, Cao, Wang, and Yang]{yang2022maniqa}
Sidi Yang, Tianhe Wu, Shuwei Shi, Shanshan Lao, Yuan Gong, Mingdeng Cao, Jiahao Wang, and Yujiu Yang.
\newblock Maniqa: Multi-dimension attention network for no-reference image quality assessment.
\newblock 2022.

\bibitem[Yue et~al.(2023)Yue, Wang, and Loy]{yue2023resshift}
Zongsheng Yue, Jianyi Wang, and Chen~Change Loy.
\newblock Resshift: Efficient diffusion model for image super-resolution by residual shifting.
\newblock \emph{Advances in Neural Information Processing Systems}, 36:\penalty0 13294--13307, 2023.

\bibitem[Zhang et~al.(2019)Zhang, Song, Gao, Chen, Bao, and Ma]{zhang2019your}
Linfeng Zhang, Jiebo Song, Anni Gao, Jingwei Chen, Chenglong Bao, and Kaisheng Ma.
\newblock Be your own teacher: Improve the performance of convolutional neural networks via self distillation.
\newblock In \emph{Proceedings of the IEEE/CVF international conference on computer vision}, pages 3713--3722, 2019.

\bibitem[Zhang et~al.(2018)Zhang, Isola, Efros, Shechtman, and Wang]{zhang2018unreasonable}
Richard Zhang, Phillip Isola, Alexei~A Efros, Eli Shechtman, and Oliver Wang.
\newblock The unreasonable effectiveness of deep features as a perceptual metric.
\newblock 2018.

\bibitem[Zhao et~al.(2022)Zhao, Cui, Song, Qiu, and Liang]{zhao2022decoupled}
Borui Zhao, Quan Cui, Renjie Song, Yiyu Qiu, and Jiajun Liang.
\newblock Decoupled knowledge distillation.
\newblock In \emph{Proceedings of the IEEE/CVF Conference on computer vision and pattern recognition}, pages 11953--11962, 2022.

\end{thebibliography}
}
\twocolumn[{%
    \begin{center}
        {\Large \textbf{\thetitle}}\\[0.5em]
        {\large Supplementary Material}\\[1.0em]
    \end{center}
}]
\renewcommand{\thesection}{\Alph{section}}
\setcounter{section}{0}

\renewcommand{\theHsection}{Supp.\thesection} 

\definecolor{suppcolor}{rgb}{0.8,0.4,0}
\newcommand{\suppref}[1]{{\color{suppcolor}#1}}

\definecolor{cvpr2}{RGB}{128,128,255}
\definecolor{TSD}{RGB}{217,174,129}
\definecolor{GAN}{RGB}{65,225,225}
\section{Implementation and Algorithm Details}

\subsection{Training Algorithm}
\label{sec:training_algo}

\begin{algorithm}[bh]
\caption{FRAMER Training Scheme}
\label{alg:framer}
\begin{algorithmic}[1]
    \footnotesize 
    
    \REQUIRE HR images $R$, Diffusion Model $\epsilon_\theta$, Teacher Layer $n$, Predefined Frequency Masks $M_{\text{LF}}, M_{\text{HF}}$ (radius $r=0.2\%$)
    \STATE \textbf{Data Preparation:}
    \STATE \hspace{1em} Generate $I_{\mathrm{LR}} \leftarrow \text{Degradation}(R)$ \COMMENT{See \refsec{sec:degradation_details}}
    \STATE \hspace{1em} Generate Caption $C \leftarrow \text{LLaVA}(I_{\mathrm{LR}})$
    \STATE \hspace{1em} Sample timestep $t \sim [1, T]$, noise $Z \sim \mathcal{N}(0, \mathbf{I})$
    \STATE \textbf{Forward Pass:}
    \STATE \hspace{1em} Encode HR image $R$ to latent $z_0$; Get noisy latent $Z_t$ using $Z, t$
    \STATE \hspace{1em} Feed $(Z_t, t, I_{\mathrm{LR}}, C)$ to $\epsilon_\theta$
    \STATE \hspace{1em} Extract Teacher Feature $\mathbf{F}^{(n)}$ and Student Features $\{\mathbf{F}^{(i)}\}_{i=1}^{N}$
    \FOR{each layer $i$ in Students}
        \STATE \textbf{Frequency Decomposition:}
        \STATE \textcolor{blue}{$\mathbf{F}^{(i)}_{\text{LF}}, \mathbf{F}^{(n)}_{\text{LF}} \leftarrow \text{FFT}(\mathbf{F}^{(i)}, \mathbf{F}^{(n)}) \odot M_{\text{LF}}$}
        \STATE \textcolor{red}{$\mathbf{F}^{(i)}_{\text{HF}}, \mathbf{F}^{(n)}_{\text{HF}} \leftarrow \text{FFT}(\mathbf{F}^{(i)}, \mathbf{F}^{(n)}) \odot M_{\text{HF}}$}
      
        \STATE \textbf{Compute Losses \& Modulation:}
        \STATE Calculate \textcolor{blue}{$\mathcal{L}^{(i)}_{\text{IntraCL}}$} using \refeq{eq:intra} \COMMENT{\textcolor{blue}{Stabilize shared structure}}
        \STATE Calculate \textcolor{red}{$\mathcal{L}^{(i)}_{\text{InterCL}}$} using \refeq{eq:inter} \COMMENT{\textcolor{red}{Sharpen instance details}}
        \STATE Compute weights $\mathbf{w}^{(i)}$ via \textbf{FAW} (\refeq{eq:faw_softmax})
        \STATE Compute alignment $a^{(i)}$ via \textbf{FAM} (\refeq{eq:fam_align})
        \STATE $\mathcal{L}^{(i)}_{\text{FRAMER}} \leftarrow$ Weighted Sum (\refeq{eq:fam_final})
    \ENDFOR
    \STATE \textbf{Optimization:}
    \STATE $\mathcal{L}_{\text{total}} = \mathcal{L}_{\text{noise}} + \sum_{i} \mathcal{L}^{(i)}_{\text{FRAMER}}$
    \STATE Update $\theta$ via Backpropagation on $\mathcal{L}_{\text{total}}$
\end{algorithmic}
\end{algorithm}

\begin{algorithm}[bh]
\caption{FRAMER Inference Scheme}
\label{alg:framer_inference}
\begin{algorithmic}[1]
    \footnotesize
    \REQUIRE LR Image $I_{\mathrm{LR}}$, Pre-trained Diffusion Model $\epsilon_\theta$
    \STATE \textbf{Preparation:}
    \STATE \hspace{1em} Generate Caption $C \leftarrow \text{LLaVA}(I_{\mathrm{LR}})$
    \STATE \hspace{1em} Sample noise $Z_T \sim \mathcal{N}(0, \mathbf{I})$ \COMMENT{Initialize at \textbf{Target HR Latent Size}}
    \STATE \textbf{Reverse Sampling Process:}
    \FOR{$t = T, \dots, 1$}
        \STATE $\epsilon_t \leftarrow \epsilon_\theta(Z_t, t, I_{\mathrm{LR}}, C)$ \COMMENT{Predict noise conditioned on LR}
        \STATE $Z_{t-1} \leftarrow \text{Sampler}(Z_t, \epsilon_t)$ \COMMENT{Denoise towards HR latent}
    \ENDFOR
    \STATE \textbf{Reconstruction:}
    \RETURN HR Image $R \leftarrow \text{Decode}(Z_0)$ \COMMENT{Maps latent to \textbf{HR Pixel Space}}
\end{algorithmic}
\end{algorithm}

\begin{table}[bh]
\centering
\caption{\textbf{Hyperparameters for the Degradation Pipeline.}}
\label{tab:degradation}
    \footnotesize 
\begin{tabular}{l|l}
\toprule
\textbf{Degradation Type} & \textbf{Parameter Settings} \\
\midrule
\multicolumn{2}{c}{\textbf{First Degradation Stage}} \\
\midrule
Blur Kernel Size & $21 \times 21$ \\
Blur Sigma & $[0.2, 3.0]$ \\
Blur Kernel Types & \makecell[l]{iso, aniso, generalized\_iso,\\generalized\_aniso, plateau\_iso,\\plateau\_aniso} \\
Sinc Probability & $0.1$ \\
Resize Range & $[0.15, 1.5]$ (Up/Down/Keep) \\
Gaussian Noise & Prob: $0.5$, Sigma: $[1, 30]$ \\
Poisson Noise & Scale: $[0.05, 3.0]$ \\
JPEG Compression & Quality: $[30, 95]$ \\
\midrule
\multicolumn{2}{c}{\textbf{Second Degradation Stage}} \\
\midrule
Blur Kernel Size & $11 \times 11$ \\
Blur Sigma & $[0.2, 1.5]$ \\
Sinc Probability & $0.1$ \\
Resize Range & $[0.3, 1.2]$ (Up/Down/Keep) \\
Gaussian Noise & Prob: $0.5$, Sigma: $[1, 25]$ \\
Poisson Noise & Scale: $[0.05, 2.5]$ \\
JPEG Compression & Quality: $[30, 95]$ \\
\midrule
\multicolumn{2}{c}{\textbf{Final Processing}} \\
\midrule
Final Sinc Prob & $0.8$ \\
Crop Size & $512$ \\
\bottomrule
\end{tabular}
\end{table}

We outline the detailed training procedure of FRAMER in \textbf{Algorithm~\ref{alg:framer}}. As described in the main paper, FRAMER is designed as a plug-and-play training strategy that leverages diffusion priors without altering inference.

\noindent\textbf{Training Phase.} 
We first synthesize Low-Resolution (LR) inputs $I_{\mathrm{LR}}$ from High-Resolution (HR) images $R$ using the Real-ESRGAN degradation pipeline~\cite{wang2021real}. Concurrently, we utilize LLaVA~\cite{liu2023llava} to generate a descriptive caption for each $I_{\mathrm{LR}}$. The diffusion model takes $(I_{\mathrm{LR}}, Z_T, \text{caption})$ as input. 
During the forward process, we extract feature maps from intermediate layers (students) and the final-layer feature map (teacher), which serves as the target representation.

As detailed in \refsec{sec:observations} of the main paper, we decompose these features into Low-Frequency (LF) and High-Frequency (HF) bands via 2D FFT using binary masks. We then compute the auxiliary distillation losses:
\begin{itemize}
    \item \textbf{IntraCL (\refsec{sec:intracl}):} Applied to the LF band to stabilize globally shared structures. It compares a student only against its teacher and a random-layer negative within the same network, avoiding false negatives common in batch-based contrastive learning.
    \item \textbf{InterCL (\refsec{sec:intercl}):} Applied to the HF band to sharpen instance-specific details. It uses in-batch negatives and random-layer negatives to promote instance discrimination and layer-wise progression.
\end{itemize}
These objectives are modulated by \textbf{FAW} (\refsec{sec:faw}), which weights distillation based on the relative frequency difference to the final layer, and gated by \textbf{FAM} (\refsec{sec:fam}), which controls distillation strength according to the student-teacher alignment. The total objective combines the standard noise-prediction loss with these frequency-aligned distillation terms (\refeq{eq:total}).

\noindent\textbf{Inference Phase.} 
We summarize the inference procedure in \textbf{Algorithm~\ref{alg:framer_inference}}. During inference, FRAMER introduces \textbf{no computational overhead}. All auxiliary heads and loss computations are strictly training-only and removed at test time. We simply generate a caption for the input LR image using LLaVA and perform standard sampling with the optimized diffusion backbone.
\subsection{Degradation Pipeline Details}
\label{sec:degradation_details}

We follow the high-order degradation process used in Real-ESRGAN~\cite{wang2021real} to synthesize training pairs. The specific parameters used in our implementation are summarized in Table~\ref{tab:degradation}.

\subsection{Computational Cost Analysis}
\label{sec:cost_analysis}

To evaluate the computational overhead introduced by FRAMER, we compare the training memory usage and time per iteration against the baseline DiT4SR model. All measurements were conducted on a single NVIDIA H200 GPU with a batch size of 16.

\begin{figure}[h]
    \centering
    \includegraphics[width=0.85\linewidth]{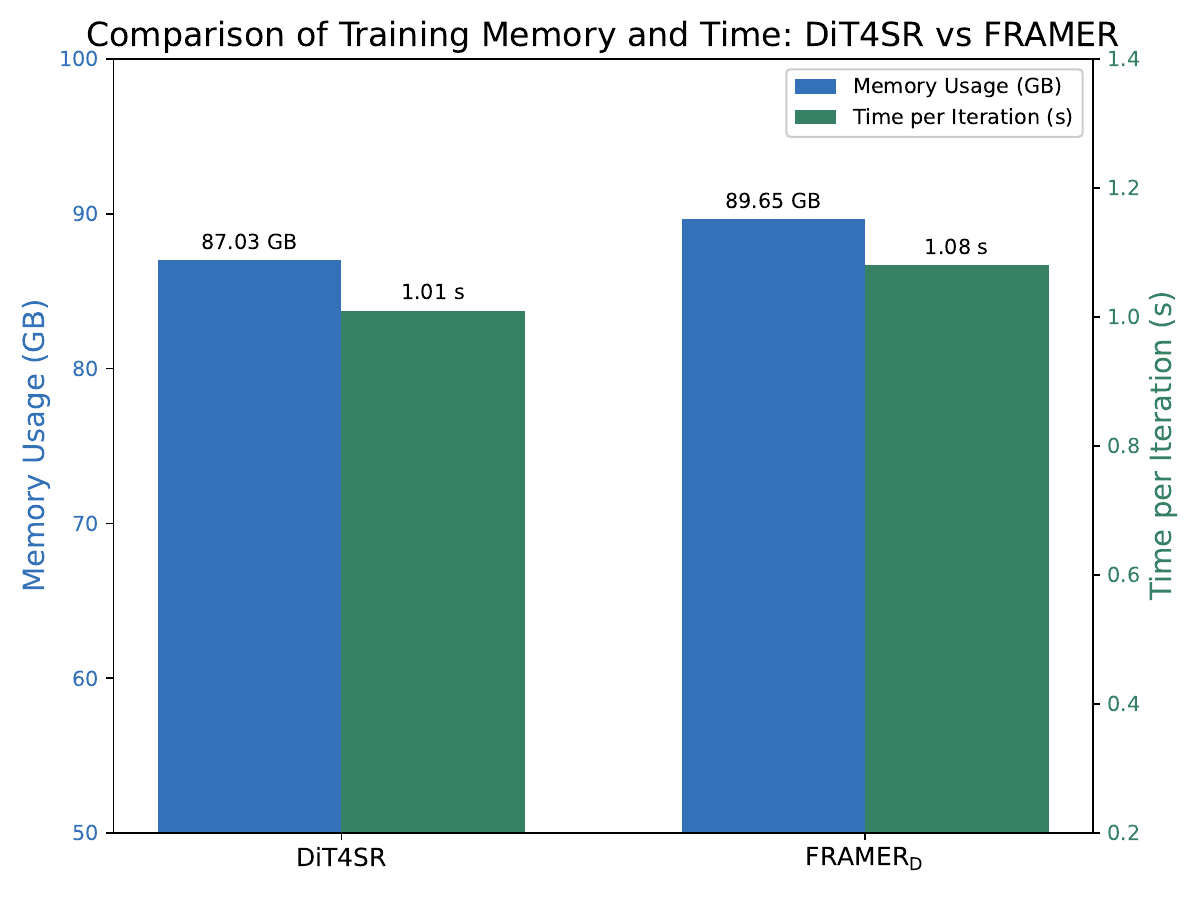}
    \caption{\textbf{Comparison of Training Cost (Memory and Time).} 
    We measure the GPU memory usage and time per iteration for DiT4SR and FRAMER$_D$ on an \textbf{NVIDIA H200 GPU with a batch size of 16}. 
    FRAMER introduces only a marginal training overhead (~3\% memory, ~7\% time) while maintaining identical inference costs due to its plug-and-play nature.}
    \label{fig:cost_analysis}
\end{figure}

As shown in \refig{fig:cost_analysis}, FRAMER incurs a marginal increase in computational resources due to the additional FFT decomposition and auxiliary loss calculations (IntraCL, InterCL). Specifically:
\begin{itemize}
    \item \textbf{Memory Usage:} Increases by approximately \textbf{3.0\%} (from 87.03 GB to 89.65 GB).
    \item \textbf{Training Time:} Increases by approximately \textbf{6.9\%} per iteration (from 1.01s to 1.08s).
\end{itemize}

Despite these slight increases during training, we consider this cost negligible given the significant improvements in convergence stability and final quality. \textbf{Crucially, FRAMER introduces zero overhead during inference.} Since the auxiliary heads and frequency-aware losses are strictly removed after training, the inference speed and memory consumption remain identical to the original backbone.

\section{Analysis of Training Dynamics and Hierarchy}

\subsection{Validation of ``Low-first, High-later'' Hierarchy}
\label{sec:supp_hierarchy}

\begin{figure}[t]
    \centering
    \includegraphics[width=1.0\linewidth]{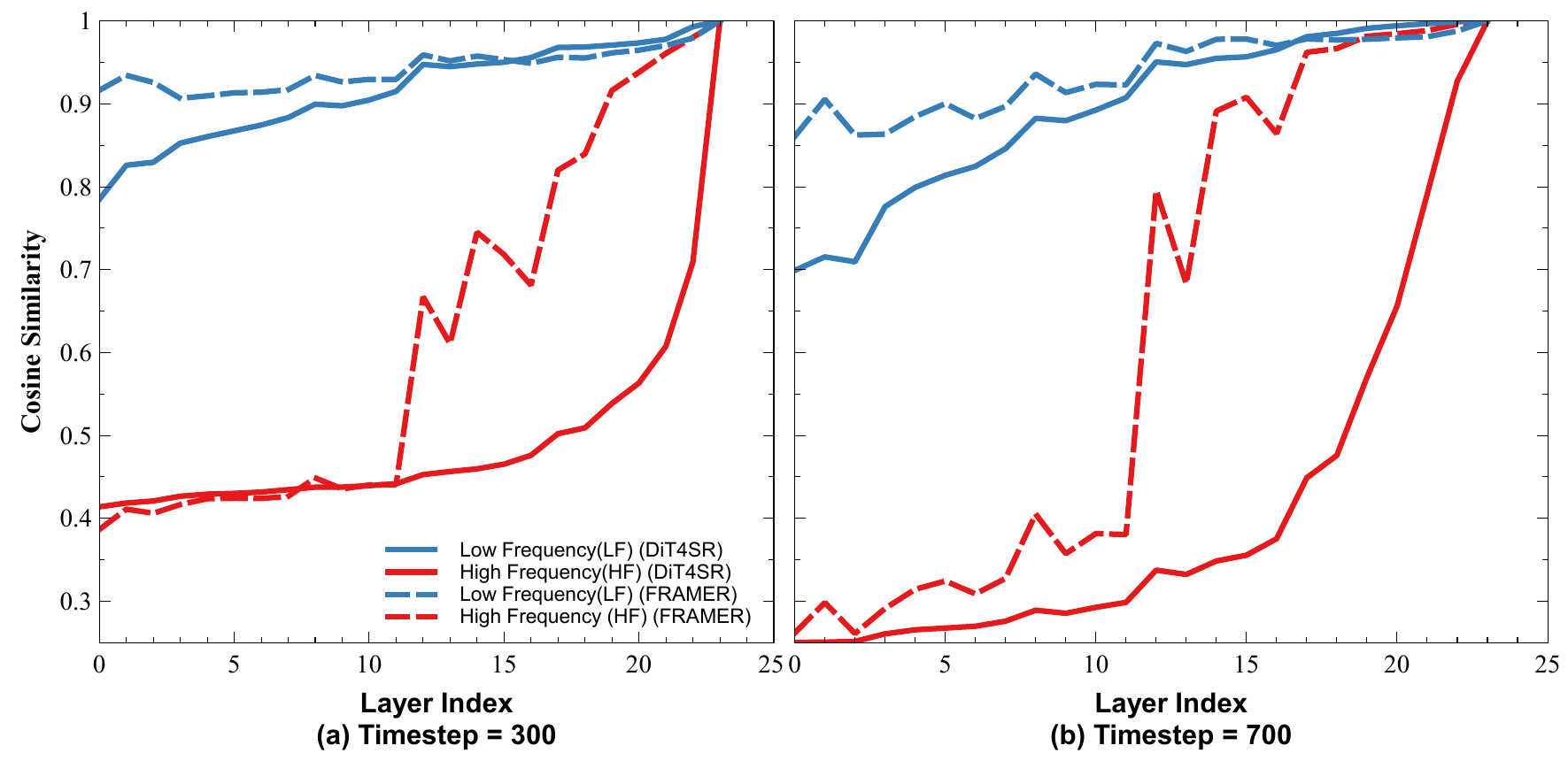}
    \vspace{-0.5cm}
    \caption{\textbf{Layer-wise cosine similarity comparison between the baseline (DiT4SR) and FRAMER.}
    We measure the similarity of intermediate features to the final-layer teacher features for \textcolor{blue}{LF (blue)} and \textcolor{red}{HF (red)} bands.
    (a) At $t=300$ and (b) $t=700$, the baseline (solid lines) shows a delayed response for HF components, validating the ``low-first, high-later'' hierarchy described in the main paper.
    In contrast, FRAMER (dashed lines) significantly accelerates HF alignment in intermediate layers (Layer 10--20), demonstrating that our frequency-aligned distillation effectively counteracts the spectral bias.}
    \label{fig:supp-cos}
\end{figure}

To empirically validate the depth-wise ``low-first, high-later'' hierarchy discussed in \refsec{sec:introduction} and \refsec{sec:observations} of the main paper, we compare the layer-wise feature alignment of our FRAMER-trained model against the baseline DiT4SR model. \refig{fig:supp-cos} plots the cosine similarity between intermediate layer features and the final-layer teacher features for both LF and HF components at two distinct noise timesteps ($t=300$ and $t=700$).

\noindent\textbf{LF Stability (Blue Lines).} As shown by the blue curves, both models achieve relatively high alignment for LF components across all layers. However, FRAMER (dashed blue) consistently maintains higher similarity scores than the baseline (solid blue), particularly in the earlier layers. This indicates that our \textbf{IntraCL} successfully stabilizes the global structure early in the network depth, preventing structural distortions during the denoising process.

\noindent\textbf{HF Acceleration (Red Lines).} The most significant difference is observed in the HF components. The baseline DiT4SR (solid red) exhibits a distinct ``low-first, high-later'' behavior: HF similarity remains near zero for the majority of the network depth and only spikes abruptly in the final few layers. This confirms our observation in the main paper that the standard noise-prediction loss leaves intermediate layers under-optimized for fine details.
In contrast, FRAMER (dashed red) demonstrates a much earlier rise in HF alignment, starting to increase significantly around Layer 10. This proves that our \textbf{InterCL}, modulated by FAW and FAM, effectively ``pre-aligns'' intermediate layers to the high-frequency details of the teacher. By mitigating the depth-wise delay in HF learning, FRAMER enables the model to dedicate more capacity to refining textures and edges throughout the network.

\subsection{Visual Analysis of Training Stability}
\label{sec:supp_convergence}

\begin{figure}[bh]
    \centering
    \includegraphics[width=1.0\linewidth]{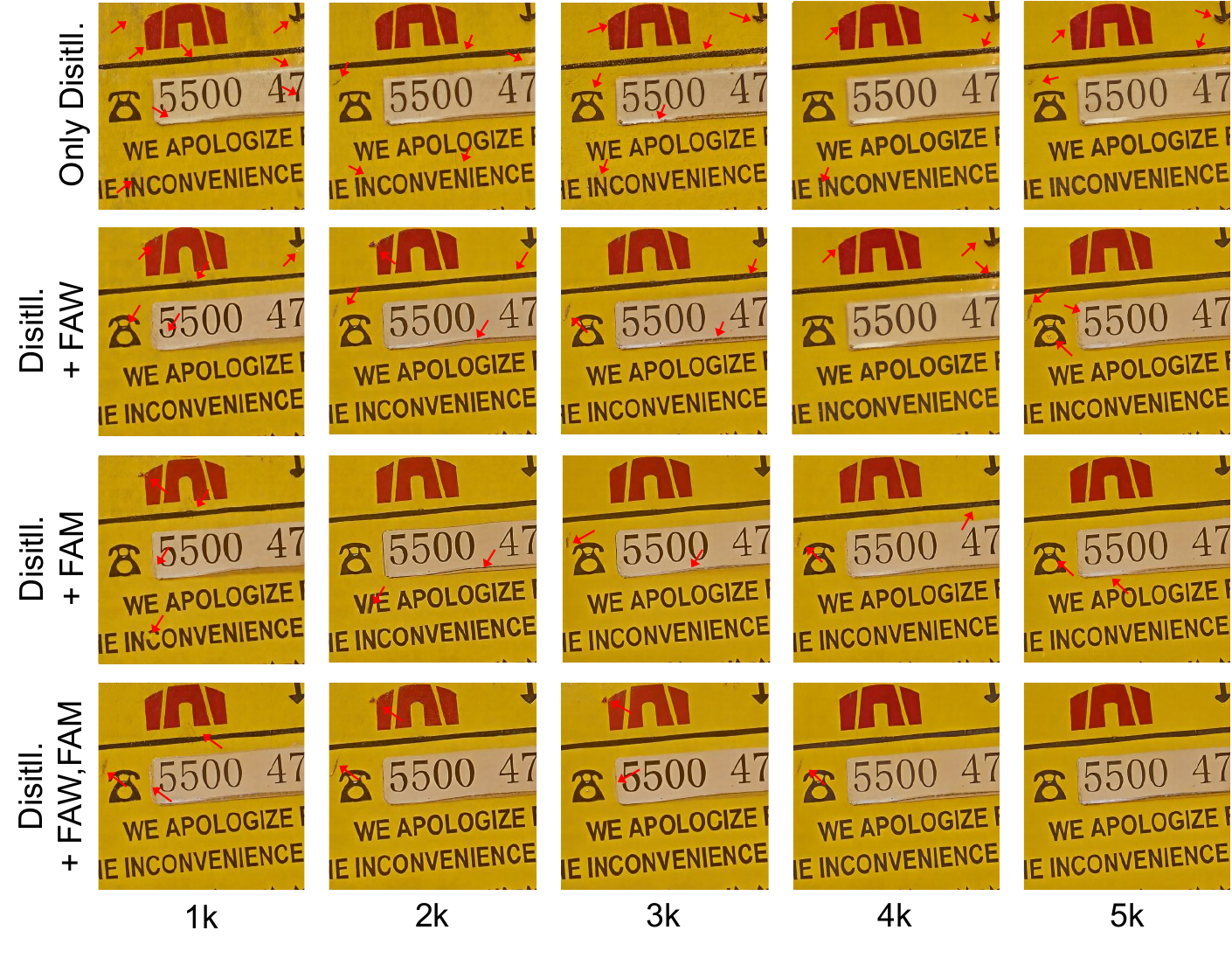}
    \vspace{-0.5cm}
    \caption{\textbf{Visual analysis of training stability during the initial phase.} 
    We compare the reconstruction quality from 1k to 5k iterations. 
    While the baseline and single-module variants show signs of instability or incoherent structures, our full method (\textbf{Distill + FAW, FAM}) demonstrates a stable optimization trajectory, effectively preventing early-stage model collapse. Red arrows indicate artifacts within each generated image. \textit{Best viewed in Zoom.}}
    \label{fig:supp_visual_convergence}
\end{figure}

To verify the effectiveness of FAW and FAM in stabilizing the optimization process and preventing potential model collapse, we conduct a visual analysis of the training dynamics during the \textbf{initial training phase} (1k--5k iterations). \refig{fig:supp_visual_convergence} compares the reconstruction quality across different configurations.

\noindent\textbf{Prevention of Early-Stage Collapse.} 
In the very early stages (1k--2k iterations), the ``Only Distill'' model and single-module variants often exhibit signs of training instability, producing chaotic artifacts or failing to form coherent structures. This suggests that aggressive self-distillation without adaptive modulation can lead to optimization difficulties or early-stage collapse, as the student is forced to mimic the teacher before establishing basic features.

\noindent\textbf{Consistent and Stable Optimization.} 
In contrast, \textbf{FRAMER (Distill + FAW, FAM)} demonstrates a stable and consistent progression throughout these initial steps. Even at 5k iterations, the model establishes a solid structural foundation with significantly reduced noise compared to other settings. This visual evidence confirms that FAM effectively gates large, unstable gradients when student-teacher alignment is low, while FAW ensures balanced frequency optimization, collectively securing a stable training trajectory from the start.

\section{Additional Analysis on Adaptive Modulation (FAW and FAM)}
\label{sec:supp_faw_fam}

In the main paper, we introduced the Frequency-based Adaptive Weight (FAW) and Frequency-based Alignment Modulation (FAM) to dynamically control the self-distillation process. In this section, we provide further empirical evidence and visualizations to support their design and complementary roles.

\subsection{FAW and FAM Weight Visualizations}
\begin{figure}[h]
\centering
\captionsetup{font=small}
\captionsetup[sub]{justification=centering}

\newcommand{\axislabelfont}{\footnotesize}

\begin{adjustbox}{width=\linewidth,center} 
\begin{tabular}{c@{\hspace{3mm}}c@{\hspace{4mm}}c}

\begin{minipage}[c]{0.7cm
}  \centering
  \rotatebox{90}{\axislabelfont Weight}
\end{minipage}
&
\begin{minipage}{3cm}
  \centering
  \includegraphics[width=\linewidth]{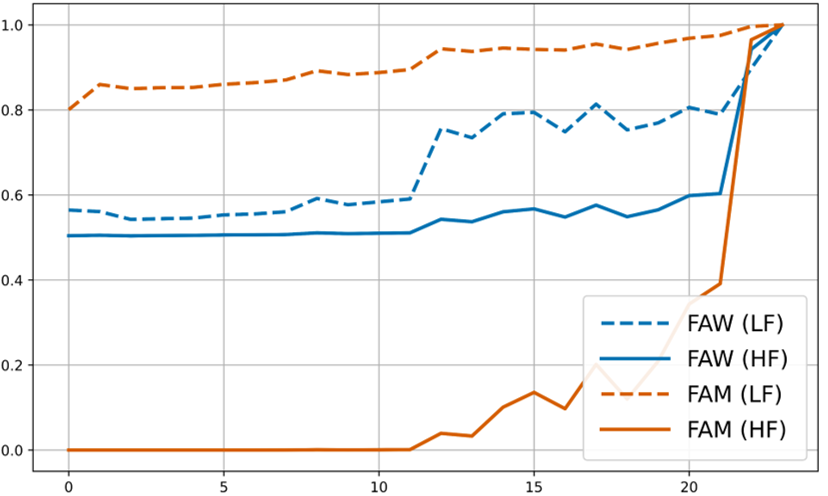}
\end{minipage}
&
\begin{minipage}{3cm}
  \centering
  \includegraphics[width=\linewidth]{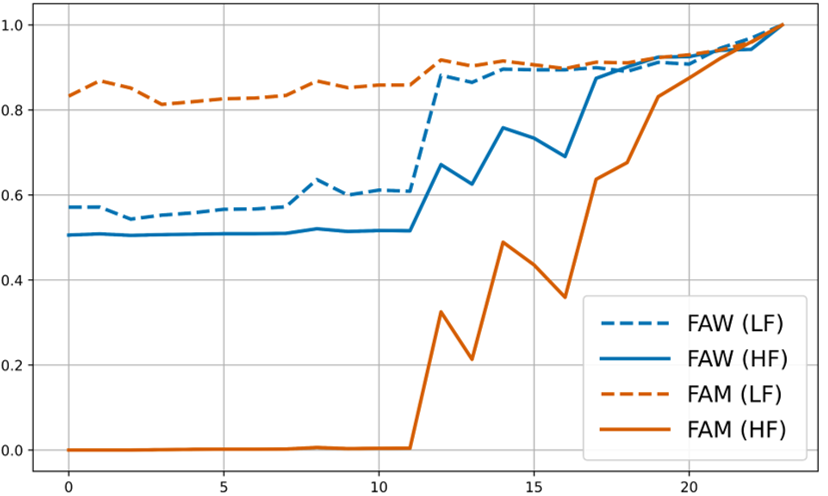}
\end{minipage}
\\[-2.0mm]

&
\multicolumn{2}{c}{
  \begin{minipage}[c]{6.4cm}
    \centering
    \axislabelfont
  \end{minipage}
}
\\[-2.0mm]

&
\begin{minipage}[t]{3cm}\centering
  \subcaption{Early training phase}
\end{minipage}
&
\begin{minipage}[t]{3cm}\centering
  \subcaption{Late training phase}
\end{minipage}

\end{tabular}
\end{adjustbox}
\caption{\textbf{Visualization of FAW/FAM weights across layers.} (a) Early training phase and (b) Late training phase. The visualizations confirm the intended behavior: HF supervision is relatively weak across most layers early on, and strengthens across more layers later in the training process.}
\label{fig:supp_weights}
\end{figure}

To demonstrate that our adaptive modules operate as intended according to the depth-wise frequency hierarchy, we visualize the layer-wise weights during different training phases in \refig{fig:supp_weights}.

As shown in the \refig{fig:supp_weights}, HF supervision is weak across most layers during the early training phase. However, as training progresses to the late phase, the HF supervision strengthens across more layers. This behavior perfectly aligns with our motivation: the network first prioritizes stabilizing the globally shared LF structures and subsequently focuses on refining the instance-specific HF details.

\subsection{Complementary Roles of FAW and FAM}
\begin{figure}[h]
\centering

\begin{adjustbox}{width=\linewidth,center} 
\setlength{\tabcolsep}{1pt}
\renewcommand{\arraystretch}{0.92}
\fontsize{5.2}{6.2}\selectfont

\newcommand{\labW}{0.95cm}
\newcommand{\colW}{1.55cm}
\newcommand{\imgW}{1.20cm}

\newcommand{\Hcell}[1]{\makebox[\colW][c]{#1}}
\newcommand{\Icell}[1]{\Hcell{\includegraphics[width=\imgW]{#1}}}

\newcolumntype{L}[1]{>{\raggedleft\arraybackslash}p{#1}} 
\newcolumntype{P}[1]{>{\arraybackslash}p{#1}}           

\begin{tabular}{@{}L{\labW} P{\colW} P{\colW} P{\colW} P{\colW}@{}}
& \Hcell{\fontsize{6}{7}\selectfont Just Distill.}
& \Hcell{\fontsize{6}{7}\selectfont \cvprref{Only FAW}}
& \Hcell{\fontsize{6}{7}\selectfont \suppref{Only FAM}}
& \Hcell{\fontsize{6}{7}\selectfont FAW\&FAM} \\[-0.1mm]

& \Icell{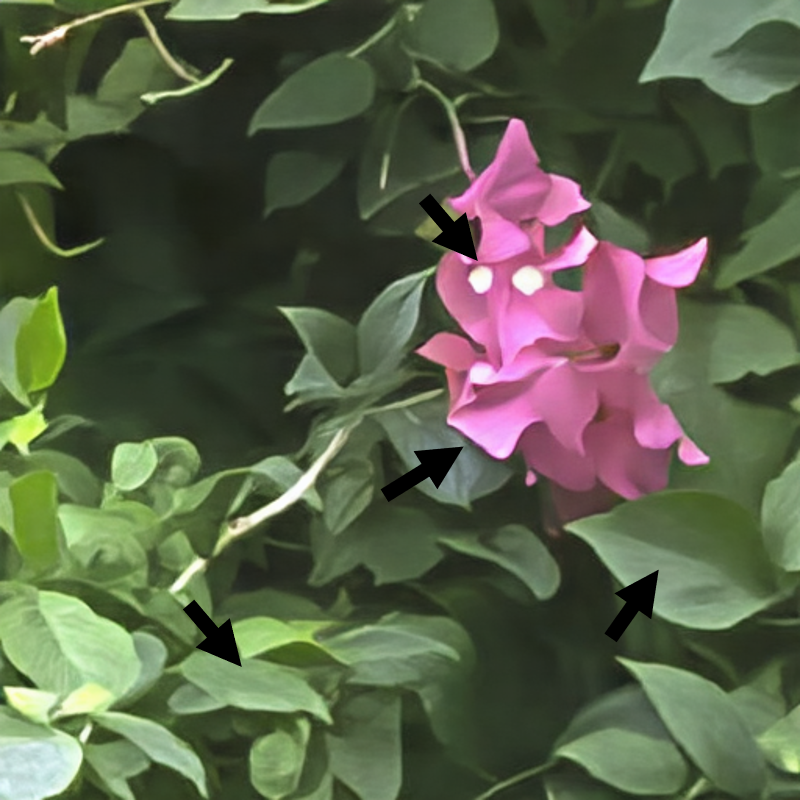} & \Icell{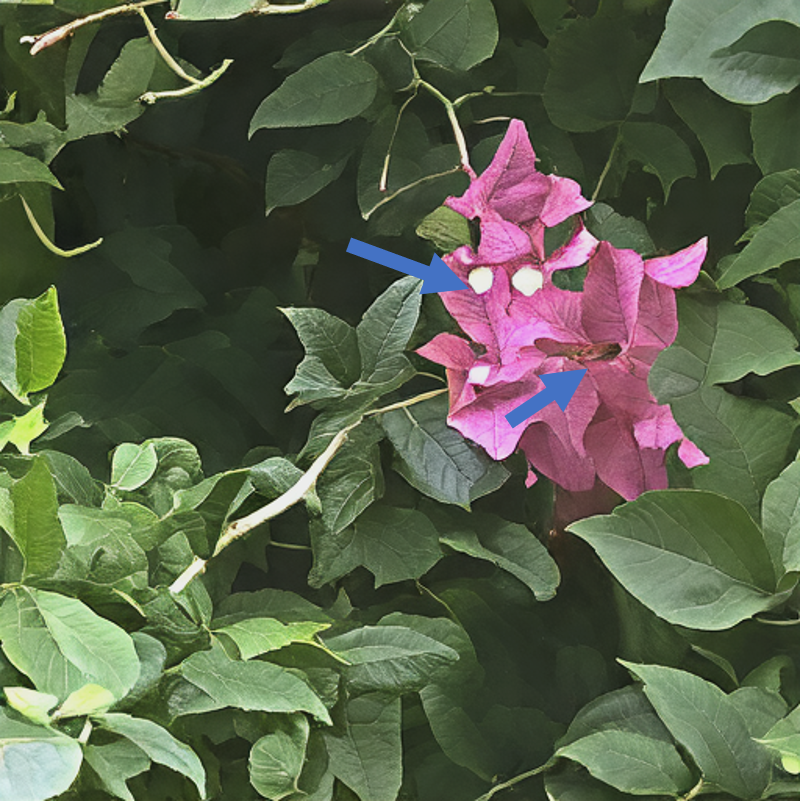} & \Icell{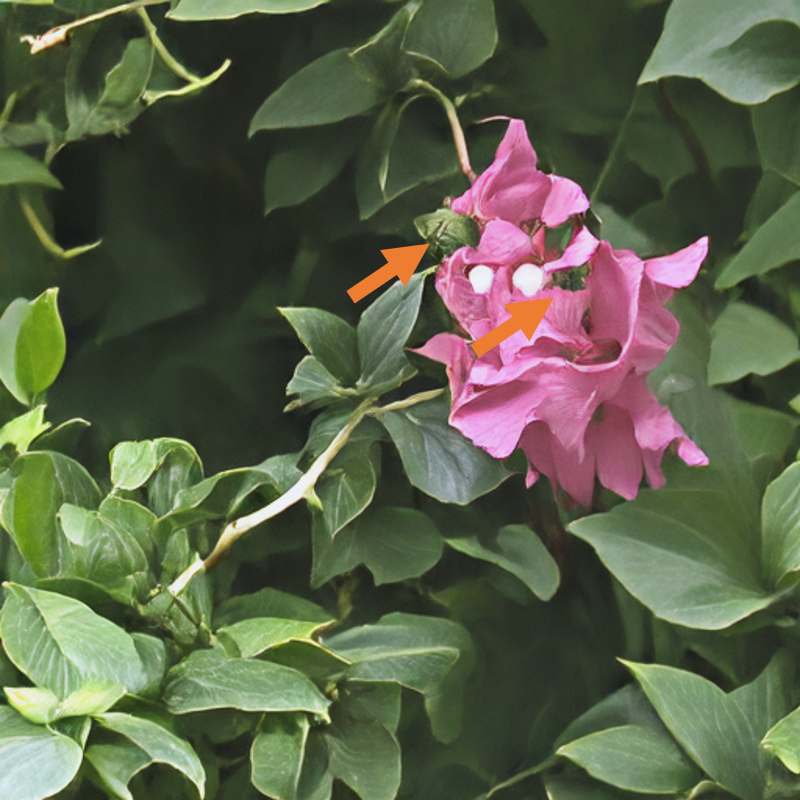} & \Icell{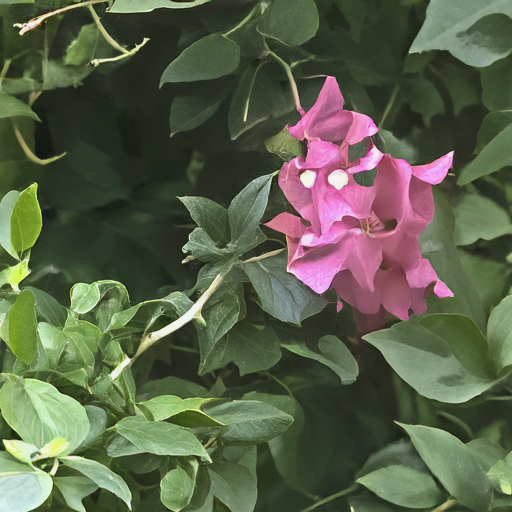} \\[-0.2mm]
NIQE$\downarrow$  & \Hcell{5.54}  & \Hcell{4.19}  & \Hcell{3.29}  & \Hcell{4.03}  \\[-0.1mm]
MANIQA$\uparrow$ & \Hcell{0.377} & \Hcell{0.540} & \Hcell{0.490} & \Hcell{0.595} \\[-0.1mm]
MUSIQ$\uparrow$  & \Hcell{67.28} & \Hcell{75.98} & \Hcell{75.71} & \Hcell{77.52} \\[0.4mm]
\\[-0.2mm]

& \Icell{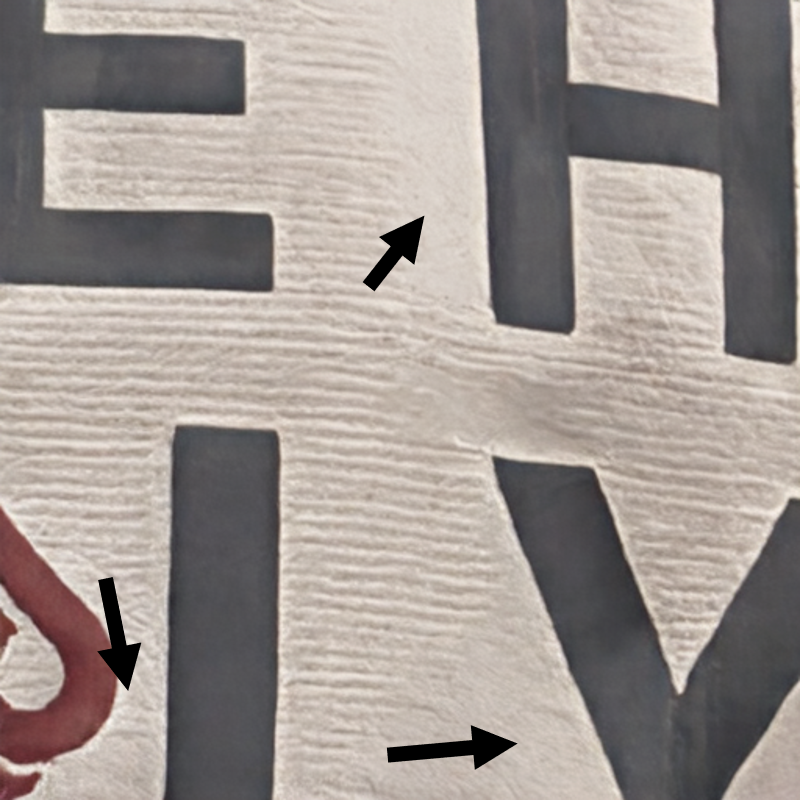} & \Icell{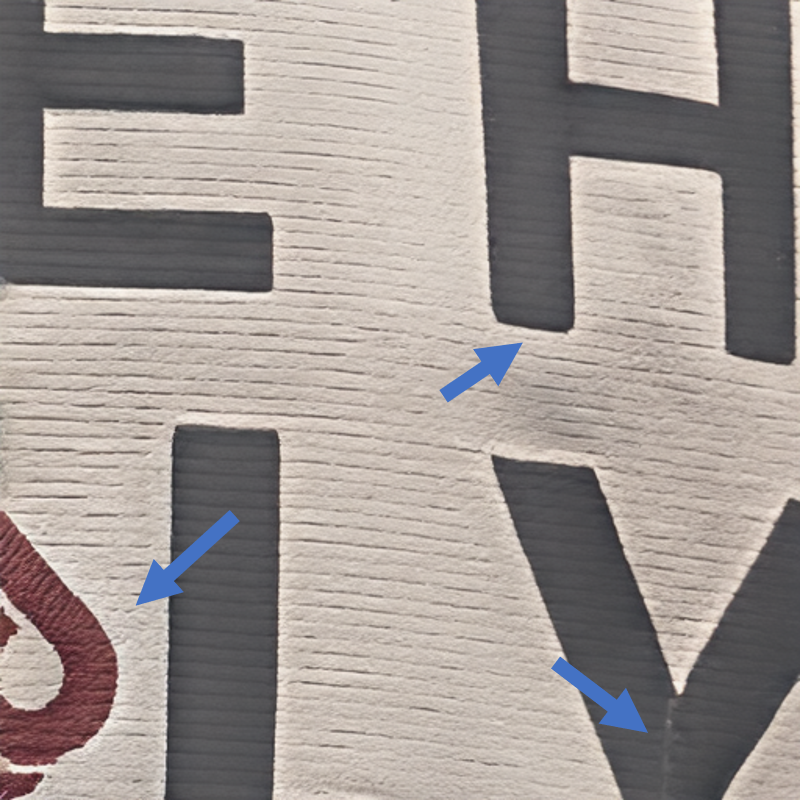} & \Icell{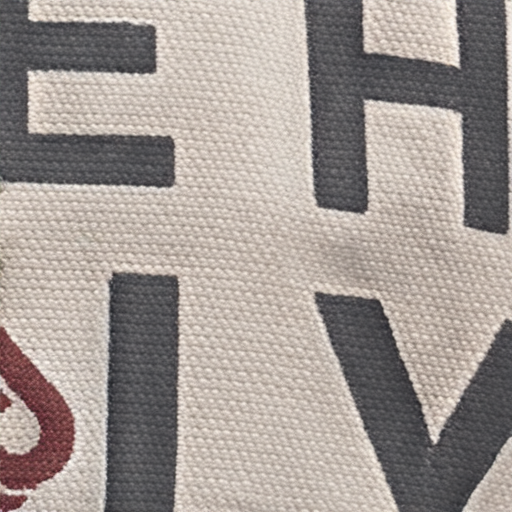} & \Icell{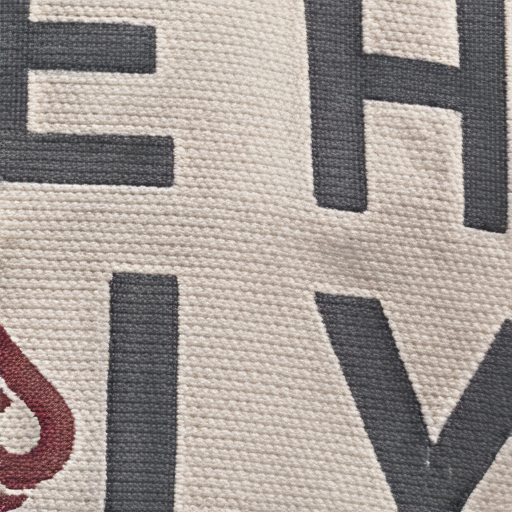} \\[-0.2mm]
NIQE$\downarrow$  & \Hcell{8.83}  & \Hcell{9.46}  & \Hcell{7.34}  & \Hcell{7.04}  \\[-0.1mm]
MANIQA$\uparrow$ & \Hcell{0.362} & \Hcell{0.396} & \Hcell{0.424} & \Hcell{0.522} \\[-0.1mm]
MUSIQ$\uparrow$  & \Hcell{57.69} & \Hcell{57.72} & \Hcell{64.37} & \Hcell{69.62} \\
\end{tabular}
\end{adjustbox}
\caption{\textbf{Qualitative comparison of adaptive modules.} Using FAW alone often causes oversharpening artifacts, while using FAM alone results in insufficient perceptual quality. FAW and FAM together produce the most coherent structures and balanced perceptual realism.}
\label{fig:supp_qualitative}
\end{figure}

We also noted the synergistic relationship between FAW and FAM. \refig{fig:supp_qualitative} provide qualitative and quantitative ablation comparisons to support this claim.

Using FAW alone rigidly scales the loss without considering the student-teacher alignment, which often leads to oversharpening artifacts. Conversely, using FAM alone lacks the frequency-specific dynamic weighting, resulting in insufficient perceptual quality. By combining both, FRAMER yields more coherent structures with balanced perceptual realism. This trend is consistently reflected in the perceptual metrics (NIQE, MANIQA, and MUSIQ), where the combined use of FAW and FAM strictly achieves the best scores.

\section{Plug-and-Play Generalization to One-step and GAN-augmented Diffusion Models}
\label{sec:supp_generalization}

In the main paper, we primarily demonstrated the effectiveness of FRAMER on multi-step diffusion backbones (U-Net and DiT). To further validate the versatility and generalization capability of our proposed framework, we extend the evaluation of FRAMER to both one-step and GAN-augmented diffusion models.

Specifically, we apply FRAMER to a representative one-step method, \textbf{TSD-SR}, and a GAN-augmented method, \textbf{SupResDiffGA}, under strictly matched training protocols. As shown in \reftbl{tab:supp_generalization}, FRAMER generalizes exceptionally well across these different model architectures and training paradigms.

On the RealSR dataset, applying FRAMER to the GAN-augmented baseline (SupResDiffGA) achieves state-of-the-art fidelity metrics (PSNR, SSIM, LPIPS), while DiT-based FRAMER models attain state-of-the-art perceptual quality. Furthermore, applying FRAMER to the one-step baseline (TSD-SR) consistently boosts both fidelity and perceptual metrics (NIQE, MANIQA, MUSIQ). Similar consistent improvements are observed on the RealLQ250 dataset. This confirms that the internal LF bias and the depth-wise frequency hierarchy are broadly shared optimization challenges in diffusion-based restoration, and FRAMER serves as an effective, backbone-agnostic solution.

\begin{table}[h]
\centering
\caption{\textbf{Quantitative comparison on one-step and GAN-augmented diffusion models.} FRAMER consistently improves both fidelity and perceptual metrics across different diffusion paradigms. The green percentages indicate the relative improvement over the respective baselines. (Note: Fidelity metrics are omitted for RealLQ250 as no ground-truth is available).}
\resizebox{\linewidth}{!}{
\begin{tabular}{l|l|ccc|ccc}
\toprule
Dataset & Model & PSNR$\uparrow$ & SSIM$\uparrow$ & LPIPS$\downarrow$ & NIQE$\downarrow$ & MANIQA$\uparrow$ & MUSIQ$\uparrow$ \\
\hline
\multirow{4}{*}{RealSR}
& \textcolor{TSD}{TSD-SR}                    
& 23.45 & 0.696 & 0.383 & 5.163 & 0.507 & 71.08 \\

& \textbf{FRAMER$_{\textcolor{TSD}{\text{TSD-SR}}}$}                   
& \textbf{24.46 (\gc{4.3\%})} 
& \textbf{0.732 (\gc{5.1\%})} 
& \textbf{0.345 (\gc{9.8\%})} 
& \textbf{4.295 (\gc{16.8\%})} 
& \textbf{0.595 (\gc{17.3\%})} 
& \textbf{73.63 (\gc{3.6\%})} \\

& \textcolor{GAN}{SupResDiffGA}               
& 24.36 & 0.697 & 0.472 & 7.294 & 0.198 & 39.57 \\

& \textbf{FRAMER$_{\textcolor{GAN}{\text{SupResDiffGA}}}$}              
& \textbf{25.72 (\gc{5.6\%})} 
& \textbf{0.751 (\gc{7.8\%})} 
& \textbf{0.416 (\gc{11.8\%})} 
& \textbf{6.229 (\gc{14.6\%})} 
& \textbf{0.230 (\gc{15.9\%})} 
& \textbf{41.59 (\gc{5.1\%})} \\
\hline
\multirow{4}{*}{RealLQ250}
& \textcolor{TSD}{TSD-SR}
& \multicolumn{3}{c|}{}
& 7.418 & 0.589 & 69.63 \\

& \textbf{FRAMER$_{\textcolor{TSD}{\text{TSD-SR}}}$}
&  &  & 
& \textbf{6.988 (\gc{5.8\%})} 
& \textbf{0.624 (\gc{6.0\%})} 
& \textbf{71.38 (\gc{2.5\%})} \\

& \textcolor{GAN}{SupResDiffGA}
&  & \textit{No ground-truth available} & 
& 5.071 & 0.343 & 57.51 \\

& \textbf{FRAMER$_{\textcolor{GAN}{\text{SupResDiffGA}}}$}    
&  &  & 
& \textbf{4.624 (\gc{8.8\%})} 
& \textbf{0.375 (\gc{9.2\%})} 
& \textbf{62.98 (\gc{9.5\%})} \\
\bottomrule
\end{tabular}
}
\label{tab:supp_generalization}
\end{table}

\section{Additional Qualitative Results}
\label{sec:supp_qualitative_results}

We provide comprehensive visual comparisons to further demonstrate the effectiveness of FRAMER. We categorize the evaluation into two groups: (1) datasets with Ground Truth (GT) references (RealSR, DrealSR) and (2) datasets without Ground Truth (RealLR200, RealLQ250), representing real-world ``in-the-wild'' scenarios.

\subsection{Comparisons on Datasets with Ground Truth}
\label{sec:supp_gt}

\begin{figure*}[t]
    \centering
    \includegraphics[width=0.7\linewidth]{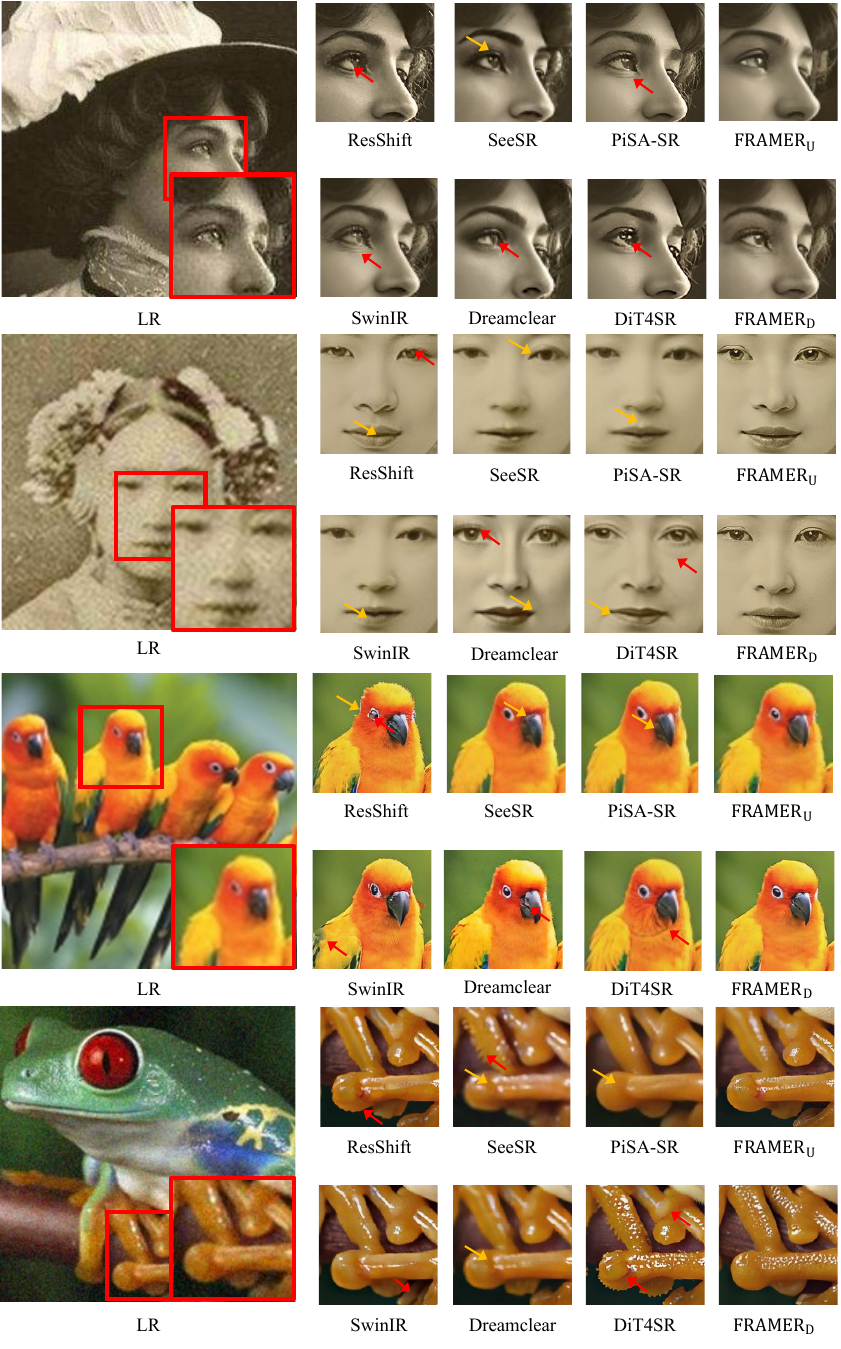}
    \caption{\textbf{Qualitative comparisons on datasets without Ground Truth (RealLR200, RealLQ250).}
    In these real-world scenarios with unknown degradations, baseline methods often suffer from severe degradations marked by arrows: \textcolor{red}{\textbf{Red}} indicates structural failures (e.g., hallucinations, object crushing), and \textcolor{yellow!90!black}{\textbf{Yellow}} indicates textural anomalies (e.g., over-sharpening, residual noise).
    FRAMER demonstrates superior perceptual quality by effectively balancing noise suppression with detail generation, avoiding these common pitfalls observed in competing methods.}
    \label{fig:supp_nongt}
\end{figure*}

\refig{fig:supp_nongt} presents comparisons on the RealLR200 and RealLQ250 datasets, which consist of low-quality real-world images with unknown degradations and no ground truth. These scenarios are particularly challenging due to severe artifacts and the risk of hallucination. Comparison methods often fail to remove heavy noise or generate unnatural artifacts (e.g., distorted facial features or blurred textures). \textbf{FRAMER} demonstrates robust generalization capabilities. For instance, in the vintage portraits, FRAMER effectively suppresses noise while enhancing facial details (eyes, hair strands) without creating uncanny artifacts. Similarly, in the animal images (parrots, frog), it restores the intricate textures of feathers and skin that are often lost by other methods. This highlights FRAMER's ability to generate perceptually pleasing and natural results even in the absence of ground truth guidance.

\section{User Study}
\label{sec:user_study}

To complement the quantitative metrics and verify the subjective superiority of our method, we conducted a user preference study.

\noindent\textbf{Experimental Setup.} We invited 15 participants to evaluate the visual quality of the restored images. The study comprised a total of 30 distinct scenes randomly selected from the RealSR and DrealSR, RealLR200, RealLQ250 datasets. Participants were asked to select the best image among the comparison methods based on three criteria: (1) \textit{Fidelity} (faithfulness to ground truth details and structure), (2) \textit{Perceptual Quality} (sharpness, naturalness, and lack of artifacts), and (3) \textit{Overall Quality} (general preference considering both fidelity and realism).

\noindent\textbf{Architecture-wise Comparison.} To ensure a rigorous and fair evaluation, we divided the study into two distinct tracks based on the backbone architecture: (1) \textbf{U-Net-based comparison} (SeeSR, PiSA-SR, vs. FRAMER$_{U}$) and (2) \textbf{DiT-based comparison} (DreamClear, DiT4SR, vs. FRAMER$_{D}$). Since different backbone architectures possess different baseline capabilities, this grouped comparison is crucial to isolate the performance gains contributed purely by our FRAMER training framework, rather than the architectural differences.

\noindent\textbf{Results and Analysis.} The results of the user study are summarized in Table~\ref{tab:user_study}. The values represent the percentage of votes where each method was selected as the best.

\begin{itemize}
    \item In the \textbf{U-Net group}, FRAMER$_{U}$ significantly outperforms the baselines, securing \textbf{56.0\%} of votes in Fidelity and \textbf{72.0\%} in Perceptual Quality. This indicates that users clearly distinguish the enhanced detail and structural stability provided by our method.
    \item In the \textbf{DiT group}, the preference for FRAMER$_{D}$ is even more pronounced, reaching \textbf{73.3\%} in Perceptual Quality. This suggests that our frequency-aligned distillation effectively unlocks the potential of the DiT backbone for generating high-frequency details that are visually pleasing to human observers.
\end{itemize}

Overall, FRAMER consistently achieves the highest preference rates across all metrics and architectures, confirming that our method produces results that are not only quantitatively superior but also perceptually more realistic and faithful.

\begin{table}[t]
    \centering
    \caption{\textbf{User Study Results (Win Rate \%).} The values indicate the percentage of votes where each method was selected as the best.
    We evaluate the win rate within each architecture group (U-Net and DiT) to ensure a fair comparison.
    \textbf{Bold} indicates the best performance.}
    \label{tab:user_study}
    \renewcommand{\arraystretch}{1.2}
    \resizebox{\linewidth}{!}{
    \begin{tabular}{l|ccc|ccc}
    \toprule
    & \multicolumn{3}{c|}{\textbf{U-Net-based Methods}} & \multicolumn{3}{c}{\textbf{DiT-based Methods}} \\
    \cmidrule(lr){2-4} \cmidrule(lr){5-7}
    \textbf{Metrics} & SeeSR~\cite{wu2024seesr} & PiSA-SR~\cite{sun2025pixel} & \textbf{FRAMER$_U$} & DreamClear~\cite{ai2024dreamclear} & DiT4SR~\cite{duan2025dit4sr} & \textbf{FRAMER$_D$} \\
    \midrule
    Fidelity ($\uparrow$)& 23.3 & 20.7 & \textbf{56.0} & 13.3 & 33.3 & \textbf{53.3} \\
    Perceptual Quality ($\uparrow$) & 12.0 & 16.0 & \textbf{72.0} & 13.3 & 13.3 & \textbf{73.3} \\
    Overall Quality ($\uparrow$) & 13.0 & 18.2 & \textbf{68.8} & 6.7 & 26.7 & \textbf{66.7} \\
        \bottomrule
    \end{tabular}
    }
\end{table}

\section{Limitations and Future Work}
\label{sec:limitations}

While FRAMER demonstrates state-of-the-art performance in restoring high-frequency details and maintaining perceptual quality, it is not entirely exempt from the inherent challenges of generative diffusion models.

\begin{figure}[t]
    \centering
    \includegraphics[width=0.9\linewidth]{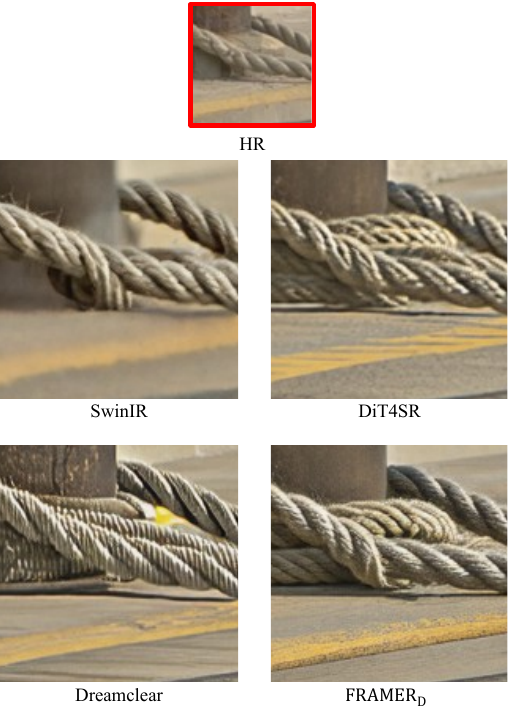}
    \caption{\textbf{Visual illustration of fidelity limitations.} 
    We compare the restoration of challenging rope textures. 
    While FRAMER$_D$ produces results that are perceptually far superior and sharper than baselines (SwinIR, DiT4SR, DreamClear), the generated fine details may exhibit slight structural deviations from the Ground Truth (HR). This illustrates the inherent trade-off between perceptual realism and pixel-wise fidelity in generative super-resolution.}
    \label{fig:supp_limitation}
\end{figure}

\noindent\textbf{Generative Artifacts.} As noted in prior studies~\cite{rombach2022high}, generative models tend to hallucinate semantic details or textures that do not exist in the original scene, especially when the input low-resolution image suffers from extreme degradation. 
Although FRAMER significantly mitigates this issue compared to pure generative baselines by enforcing feature consistency via self-distillation, minor artifacts or unfaithful texture synthesis may still occur. 
For instance, as shown in \refig{fig:supp_limitation}, while our method reconstructs the rope texture with high definition compared to the blurry or artifact-prone baselines, the specific twisting pattern may slightly diverge from the exact pixel-structure of the Ground Truth.

\noindent\textbf{Future Directions.} To further address the stochastic nature of diffusion-based upscaling and minimize hallucinations, future work could explore integrating frequency-constrained sampling strategies. For instance, adapting training-free inference techniques like \textbf{FouriScale}~\cite{lin2024fouriscale}, which manipulates frequency components during the reverse sampling process to ensure structural rigidity, could complement our training-time frequency alignment. Combining FRAMER's robust representation learning with such inference-time constraints represents a promising avenue for achieving hallucination-free, high-fidelity super-resolution.

\begin{figure*}[t]
    \centering
    \includegraphics[width=0.68\linewidth]{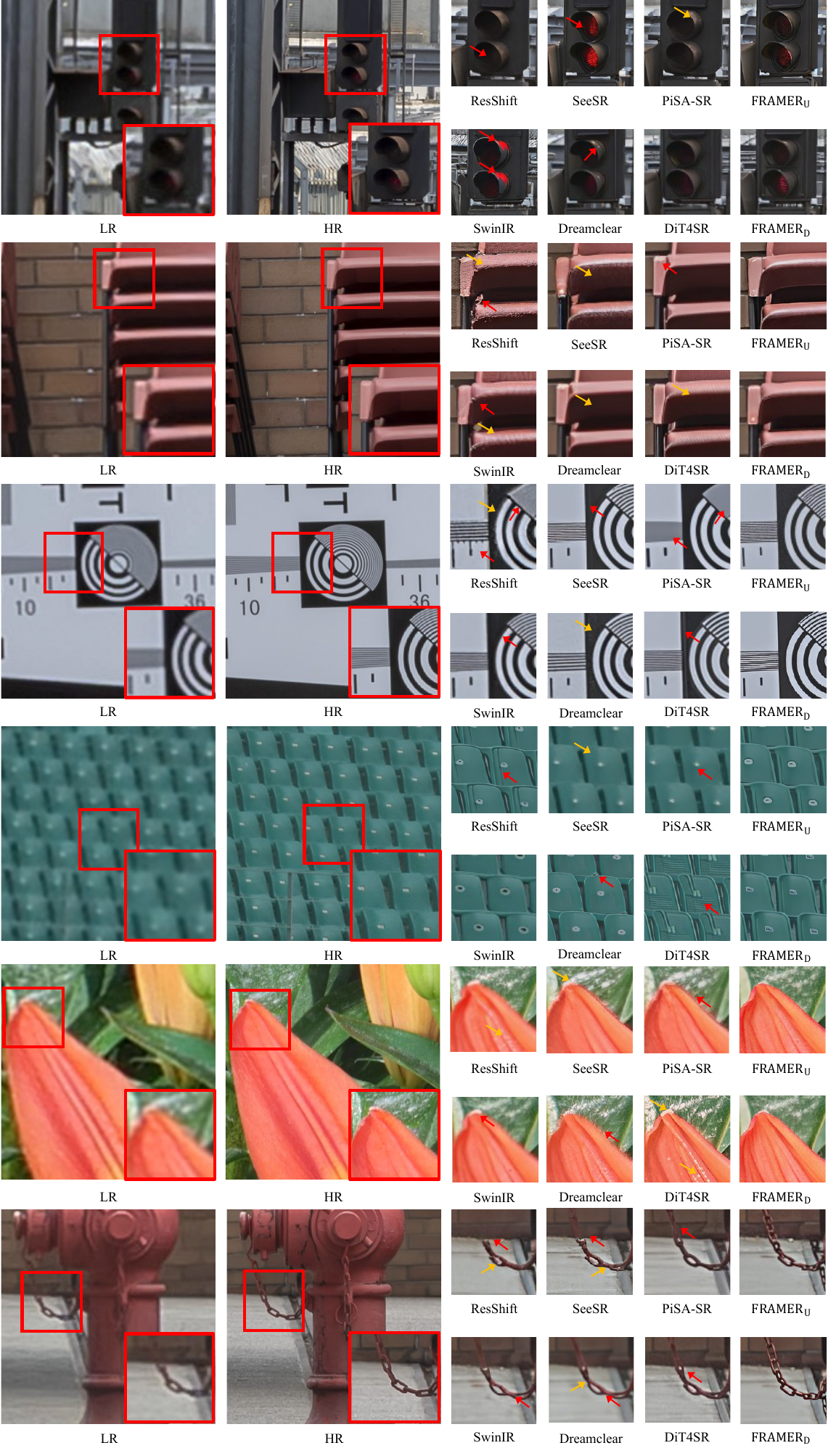}
    \caption{\textbf{Qualitative comparisons on datasets with Ground Truth (RealSR, DrealSR).}
    We compare FRAMER against state-of-the-art methods (SwinIR, ResShift, SeeSR, PiSA-SR, DreamClear, DiT4SR).
    We highlight specific failure cases in baseline methods: \textcolor{red}{\textbf{Red arrows}} indicate structural errors (e.g., hallucinations, object distortion), while \textcolor{yellow!90!black}{\textbf{Yellow arrows}} point to textural defects (e.g., over-sharpening, blur, noise).
    In contrast, our methods (FRAMER$_{U}$, FRAMER$_{D}$) consistently mitigate these artifacts, producing sharper edges and faithful textures closely aligning with the HR references.}
    \label{fig:supp_gt}
\end{figure*}

\end{document}